\documentclass{article}

\usepackage[final]{neurips_2024}

\usepackage{todonotes}

\usepackage[utf8]{inputenc} % allow utf-8 input
\usepackage[T1]{fontenc}    % use 8-bit T1 fonts
\usepackage{hyperref}       % hyperlinks
\usepackage{url}            % simple URL typesetting
\usepackage[ruled, noend]{algorithm2e}
\usepackage{enumitem}
\usepackage{booktabs}       % professional-quality tables
\usepackage{amsfonts}       % blackboard math symbols
\usepackage{nicefrac}       % compact symbols for 1/2, etc.
\usepackage{microtype}      % microtypography
\usepackage{xcolor}         % colors
\usepackage{graphicx}
\usepackage{amsmath}
\usepackage{subcaption}
\usepackage{multirow}

\usepackage{xspace}
\newcommand{\method}{DiTFastAttn\xspace}
\newcommand{\warsshort}{WA-RS}

\newcommand{\assshort}{AST}
\newcommand{\ascshort}{ASC}
\newcommand{\wars}{Window Attention with Residual Sharing}
\newcommand{\wa}{Window Attention}
\newcommand{\ass}{Attention Sharing across Timesteps}
\newcommand{\asc}{Attention Sharing across CFG}

\title{\method{}: Attention Compression for Diffusion Transformer Models}

\author{
    Zhihang Yuan\thanks{Equal contribution. $^\dag$ Corresponding author and Project advisor. Project Website: \url{http://nics-effalg.com/DiTFastAttn}.}~~$^{1,2}$
    \And
    Hanling Zhang$^{*1,2}$
    \And
    Pu Lu$^{*1}$
    \And
    Xuefei Ning$^{1~\dag}$
    \And
    Linfeng Zhang$^{3}$
    \And
    Tianchen Zhao$^{1,2}$
    \And
    Shengen Yan$^{2}$
    \And
    Guohao Dai$^{3,2}$
    \And
    Yu Wang$^{1~\dag}$
}

\begin{document}

\maketitle

\vspace{-10pt}

\begin{center}
{\fontsize{8.5pt}{15pt}\selectfont
$^1$Tsinghua University~~~~$^2$Infinigence AI~~~~$^3$Shanghai Jiao Tong University
}
\vspace{10pt}
\end{center}

\begin{abstract}
Diffusion Transformers (DiT) excel at image and video generation but face computational challenges due to the quadratic complexity of self-attention operators. We propose \method{}, a  post-training compression method to alleviate the computational bottleneck of DiT.
We identify three key redundancies in the attention computation during DiT inference: (1) spatial redundancy, where many attention heads focus on local information; (2) temporal redundancy, with high similarity between the attention outputs of neighboring steps; (3) conditional redundancy, where conditional and unconditional inferences exhibit significant similarity. We propose three techniques to reduce these redundancies: (1) \textit{Window Attention with Residual Sharing} to reduce spatial redundancy; (2) \textit{Attention Sharing across Timesteps} to exploit the similarity between steps; (3) \textit{Attention Sharing across CFG} to skip redundant computations during conditional generation. %Our method compresses the model FLOPs to 35\% of the original model. %This work offers a practical solution for deploying DiT models in resource-constrained environments.
% 1
We apply \method{} to DiT, PixArt-Sigma for image generation tasks, and OpenSora for video generation tasks. Our results show that for image generation, our method reduces up to 76\% of the attention FLOPs and achieves up to 1.8$\times$ end-to-end speedup at high-resolution (2k $\times$ 2k) generation. %For video generation with OpenSora, our method compresses the model FLOPs to 48\% of the original model at 240p resolution with 16 frames.
% 2
%Experimental results show that \method{} consistently reduces computation costs on DiT models including DiT-XL, PixArt-Sigma, and Open-Sora. \method{} achieves a 36\% to 88\% attention computation reduction and up to 1.6x speedup on 2K image generation task.

\end{abstract}

\section{Introduction}

\begin{figure}

\begin{minipage}{\textwidth}
    \begin{subfigure}{0.36\textwidth}
    \centering
    \includegraphics[width=\textwidth]{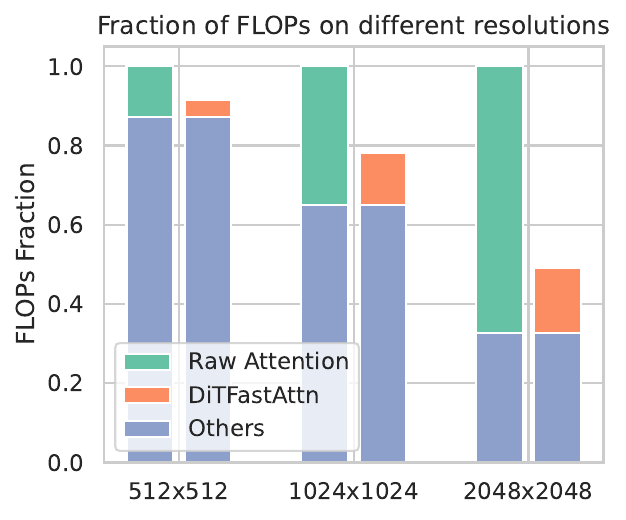}
    % \caption{Summarized Performance of \asvd.}
    % \label{fig:performance-summary}
    \end{subfigure}
    \hfill
    \begin{subfigure}{0.62\textwidth}
    \centering

    \includegraphics[width=\textwidth]{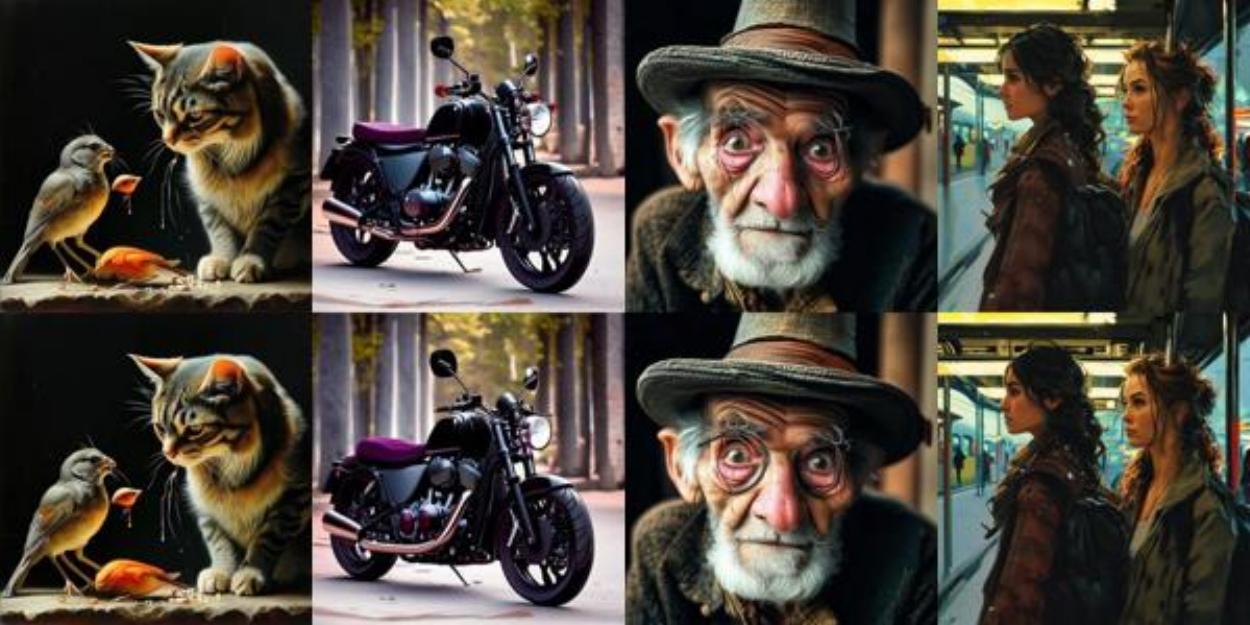}
    % \caption{High-level idea of using \textit{\asvd~to compress KV cache}.}
    % \label{fig:kvcompression-idea}
    \end{subfigure}
    \end{minipage}
\caption{\textbf{Left}: The efficiency benefits of applying \method{} on PixArt-Sigma~\citep{chen2024pixart} when generating images of different resolutions. The Y-axis shows the \#FLOPs fraction normalized by the \#FLOPs of the original model. \textbf{Right}: The qualitative results of applying \method{} on 1024$\times$1024 PixArt-Sigma.}
\label{fig:intro}
\end{figure}

Recently, diffusion transformers (DiT) have gained increasing popularity in image~\citep{peebles2023scalable,chen2024pixart} and video generation~\citep{videoworldsimulators2024}. %Diffusion transformers have facilitated many applications, including image editing, text-to-image generation, and video generation.
% Given the increasing popularity of Diffusion Transformer (DiT) in image and video generation, attention optimization has become crucial. 
However, a major challenge with DiTs is their substantial computational demand, particularly noticeable when generating high-resolution content. On the one hand, traditional transformer architectures, with their self-attention mechanism, have an $\mathcal{O}(L^2)$ complexity to the input token length $L$. This quadratic complexity leads to significant computational costs as the resolution of images and videos escalates. As demonstrated in Fig.\ref{fig:intro}, the attention computation becomes the primary computational bottleneck during the inference process as image resolution increases. 
%Specifically, tokenizing a \(2k \times 2k\) image into 16,000 tokens necessitates several seconds for attention computation when executed on high-end GPU servers such as the Nvidia A100.
Specifically, if a \(2K \times 2K\) image is tokenized into 16k tokens~\citep{chen2024pixart}, this requires several seconds for attention computation, even on high-end GPUs such as the Nvidia A100.
On the other hand, the inference process of diffusion requires a substantial number of neural network inferences due to the multiple denoising steps and the classifier-free guidance (CFG) technique~\citep{ho2022classifier}.
%Therefore, the optimization of the attention computation is crucial for the application of DiT models.
%in the Pixart-sigma model, the resolution increase of generated images leads to attention computation becoming the main computational burden during the inference process. 
% For example, in Pixart-alpha, a \( 2k \times 2k \) resolution image results in 16k tokens, demanding several seconds even on high-end GPU servers such as Nvidia A100-40G for self-attention computation. 
%Therefore, the optimization of DiT inference has become crucial.

Previous efforts to accelerate attention mechanisms, such as Local Attention, Swin Transformer~\citep{liu2021swin}, and Group Query Attention (GQA)~\citep{ainslie2023gqa}, mainly focused on designing the attention mechanism 
%redesign 
% attention sharing
or network architecture. % redesign.
While effective in reducing computational costs, these approaches necessitate large retraining costs. Due to the substantial data and computational requirements for training a DiT, there is a need for post-training compression methods.
In this work, we identify three types of redundancy in the attention computation of DiT inference and propose a post-training model compression method, \method{}, to address these redundancies:

(1) \textit{Redundancy in the spatial dimension}. Many attention heads primarily capture local spatial information, with attention values for distant tokens nearing zero. %As shown in Fig. \ref{fig:residual_share}, attention values of some heads are concentrated within a diagonal window. 
To reduce the redundancy, we opt to use window attention instead of full attention for certain layers. However, directly discarding all attention computation outside the window leads to significant performance degradation. To maintain the performance in a training-free way, we propose to cache the residual between the outputs of full and window attention at one step and reuse this residual for several subsequent steps. We refer to this technique as \textbf{\wars{} (\warsshort{})}.
%To reduce spatial dimension redundancy, we propose using window attention to replace full attention. Additionally, Fig. \ref{fig:residual_share} shows that values inside the window change significantly while those outside change minimally. We propose caching the minimally changing values outside the window as residuals, which can be reused in subsequent steps to maintain generation performance.

(2) \textit{Similarity between the neighboring steps in attention outputs}. 
The attention outputs of the same attention head across neighboring steps can be highly similar. We propose the \textbf{\ass{} (\assshort{})} technique that exploits this step-wise similarity to accelerate attention computation.
%Second, in DiT, we observe similarity across neighboring steps in attention outputs. By exploiting this step-wise similarity, we introduce attention sharing across step to reduce attention computation.

(3) \textit{Similarity between the conditional and unconditional inference in attention outputs}. We observe that in CFG, the attention outputs of conditional and unconditional inference exhibit significant similarity (SSIM $\geq 0.95$) for certain heads and timesteps. Therefore, we propose the \textbf{\asc{} (\ascshort{})} technique to skip redundant computation during unconditional generation.
% To enhance computational efficiency, DiTFastAttention reorders attention heads within each layer of the Transformer block based on their attention pattern, consolidating similar heads. This transformation optimizes kernel launches, improving computational efficiency. Furthermore, we address practical deployment scenarios by offering two execution modes: Dynamic mode, where attention scales are computed on-the-fly, and Static mode, where attention scales for each head are fixed during calibration, ensuring predictable latency.

% \begin{figure}[t]
%     \centering
%     \includegraphics[width=0.96\textwidth]{images/intro.png}
%     \caption{\textbf{Left}: The efficiency benefits of applying \method{} on PixArt-Sigma~\citep{chen2024pixart} when generating images of different resolutions. The Y-axis shows the \#FLOPs fraction normalized by the \#FLOPs of the original model. \textbf{Right}: The qualitative results of applying \method{} on 1024$\times$1024 PixArt-Sigma.}
%     \label{fig:intro}
% \end{figure}

We conduct extensive experiments to evaluate \method{} using multiple DiT models, including DiT-XL~\citep{peebles2023scalable} and PixArt-Sigma~\citep{chen2024pixart} for image generation, and Open-Sora~\citep{Open-Sora} for video generation.
Our findings demonstrate that \method{} consistently reduces the computational cost. Notably, the higher the resolution, the greater the savings in computation and latency. For instance, with PixArt-Sigma, \method{} delivers a 20\% to 76\% reduction in attention computation and a end-to-end speedup of up to 1.8x during the generation of 2048$\times$2048 images.
% The experiments shows that \method{} can consistently reduce the computation cost reduce the inference latency. The larger the resolution, the higher reduction our method can achieve.
% For instance, \method{} achieves 36\% to 88\% attention computation reduction and up to 37\% of latency reduction on 2048x2048 image generation using PixArt-Sigma.
% \method{} can compress the model FLOPS to 35.54\% of the original and reduce latency to 76.32\% of the original, while maintaining the model's generative performance.
% \xuefei{be more specific on one highlight number, or write a range; also give some results on latency?}
% we conduct extensive experiments to evaluate DiTFastAttention with multiple diffusion transformers(DiT-XL,Pixart-Sigma,Open-Sora) om ImageNet and COCO. Our empirical results demanstrate that DiTFastAttention can compress Flops to 35.54\% with 64.27\% latency while maintaining the generation quality.
% As shown in Fig.\ref{fig:intro}, for the 2K image generation task, the application of DiTFastAttn reduces attention calculation by 70\%, thereby resulting in a 2x speed while maintaining the generation quality.
% As shown in Fig.\ref{fig:intro}, by applying DiTFastAttn,  70\% of attention calculation is reduced, resulting in a 2x speed for the 2K image generation task while maintaining the generation 质量.

%Experimental results demonstrate that under static model constraints, DiT model achieves 1.1x - 1.4x acceleration for generating \( 1K \times 1K \) images and 1.2x - 1.5x for \( 2K \times 2K \) images. 

\section{Related Work}

\subsection{Diffusion Models}
Diffusion models~\citep{ho2020denoising,rombach2022high, peebles2023scalable, chen2024pixart, videoworldsimulators2024} have gained significant attention due to their superior generative performance compared to GANs~\citep{creswell2018generative}.
% Denoising diffusion probabilistic models (DDPMs)~\citep{ho2020denoising}, operating on the principle of denoising process to transform Gaussian noise into an image, have emerged as a successful approach and have been widely used in image and video generation tasks.
% Latent diffusion models (LDMs)~\citep{rombach2022high} improve diffusion models' training efficiency and conditioning with latent variable model.
Early diffusion models~\citep{ho2020denoising,rombach2022high} are implemented based on the U-Net architecture. 
To achieve better scalability, DiT \citep{peebles2023scalable} utilizes the transformer architecture instead of U-Net.
% Recent notable contributions of diffusion models include 
Diffusion transformer is applied in the fields of image and video generation.
PixArt-Sigma~\citep{chen2024pixart} demonstrates the diffusion transformer's capability to generate high-resolution images up to 4K. Sora~\citep{videoworldsimulators2024} presents the diffusion transformer's ability to generate videos.

\subsection{Vision Transformer Compression}
The computational overhead of attention has garnered significant attention. FlashAttention~\citep{dao2023flashattention} divides the input tokens into smaller tiles to minimize redundant memory accesses and optimize latency.
Some studies highlight the quadratic complexity of attention computation and improve efficiency through token pruning, achieved by filtering~\citep{rao2021dynamicvit,liu2022adaptive,wu2023ppt} or merging~\citep{lu2023content, huang2023vision,wu2023ppt} tokens at different stages of the network.
DynamicViT~\citep{rao2021dynamicvit} employs a prediction network to dynamically filter tokens.
Adaptive Sparse ViT~\citep{liu2022adaptive} filters tokens by simultaneously considering the attention values and the L2 norm of the features.
~\cite{lu2023content} trains a network with segmentation labels to direct the merging operations of tokens in regions with similar content.
~\cite{huang2023vision} conducts attention computations after downsampling tokens and subsequently upsampling to recover the spatial resolution.
~\cite{wu2023ppt} demonstrates that deeper layers are more suitable for filtering tokens, whereas shallower layers are more appropriate for merging tokens.
\subsection{Local Attention}

Various studies have delved into the utilization of local attention patterns, where each token attends to a set of neighboring tokens within a fixed window size, aiming to mitigate the computational burden associated with processing long sequences. The concept of local windowed attention was initially introduced by ~\cite{beltagylongformer} in Longformer, presenting an attention mechanism that scales linearly with sequence length. Bigbird~\citep{zaheer2020big} extends this idea by incorporating window attention, random attention, and global attention mechanisms, enabling the retention of long-range dependencies while mitigating computational costs. In the realm of computer vision, Swin Transformer~\citep{liu2021swin} adopts a similar approach by confining attention computation to non-overlapping local windows, utilizing shifted windows across different layers to capture global context efficiently. Twins Transformer\citep{chu2021twins}, FasterViT\citep{vasu2023fastvit}, and Neighborhood attention transformer \citep{hassani2023neighborhood} employ window-based attention to enhance computational efficiency, leveraging different module designs such as global sub-sampled attention and hierarchical attention to exploit global context effectively. In our work, we employ fixed-sized window attention to accelerate pretrained Diffusion Transformer models and introduce a novel technique named \wars{} to preserve long-range dependencies for image tokens.

\subsection{Attention Sharing}

% GQA divides the query heads into \( G \) groups. Each query retains its own parameters, while each group shares a key and value, reducing memory read and improving efficiency. PSVIT shows that attention maps between different layers in ViT have significant similarity and suggests sharing attention maps across layers to reduce redundant computation. Deepcache demonstrates that the high-level features of the UNET framework diffusion models show significant similarity across timesteps. They propose two methods to reuse UNET's high-level features, skipping intermediate layers' computation to accelerate the denoising process. TGATE shows that the cross-attention output of text-conditional diffusion models converges to a fixed point after several inference steps. TGATE caches the cross-attention output once it converges and keeps it fixed during the remaining denoising steps to reduce computation.

GQA~\citep{ainslie2023gqa} divides query heads into \( G \) groups. Each query retains its own parameters, while each group shares a key and value, reducing memory usage and improving efficiency. PSVIT~\citep{chen2021psvit} shows that attention maps between different layers in ViT have significant similarity and suggests sharing attention maps across layers to reduce redundant computation. Deepcache~\citep{ma2023deepcache} demonstrates that high-level features in U-Net framework diffusion models are similar across timesteps. Deepcache proposes reusing U-Net's high-level features and skipping intermediate layers' computation to accelerate denoising process. TGATE~\citep{zhang2024cross} shows that the cross-attention output of text-conditional diffusion models converges to a fixed point after several denoising timesteps. TGATE caches this output once it converges and keeps it fixed during the remaining denoising steps to reduce computational cost. In \method{}, we demonstrate the similarity of attention outputs both CFG-wise and step-wise. We also consider the differences in similarity across different layers at various steps to share attention outputs CFG-wise and step-wise.

\subsection{Other Methods to Accelerate Diffusion Models}

Network quantization is a widely used technique for reducing the bitwidth of weights and activations, effectively compressing both image generation models \citep{shang2023post,zhao2024mixdq} and video generation models \citep{zhao2024vidit}.
Scheduler optimization is another popular approach aimed at decreasing the number of timesteps in the denoising process \citep{song2020denoising,lu2022dpm,liu2023oms}. Additionally, distillation serves as an effective method for minimizing the timesteps required during denoising \citep{DBLP:conf/iclr/SalimansH22,meng2023distillation,liu2023instaflow}.
DiTFastAttn offers a complementary solution, as it operates independently of the specific quantization bitwidth, scheduler, and timesteps employed.

% Network quantization is a widely used methods to reduce the bitwidth of the weights and activations, which is used to compress both image generation models~\cite{shang2023post,zhao2024mixdq} and video generation models~\cite{zhao2024vidit}. 
% Scheduler optimization is another popular method to reduce the timesteps in denoising process~\cite{song2020denoising,lu2022dpm,liu2023oms}.
% Distillation is anther method to reduce the timestep required in denoising~\cite{DBLP:conf/iclr/SalimansH22,meng2023distillation,liu2023instaflow}.
% DiTFastAttn is orthogonal to these methods that it can work in any given quantization bitwith, scheduler and timesteps.

\section{Method}

\subsection{Overview}
% A table for different dimension sharing
% \usepackage{booktabs}

\begin{figure}[t]
    \centering
    \includegraphics[width=0.95\textwidth]{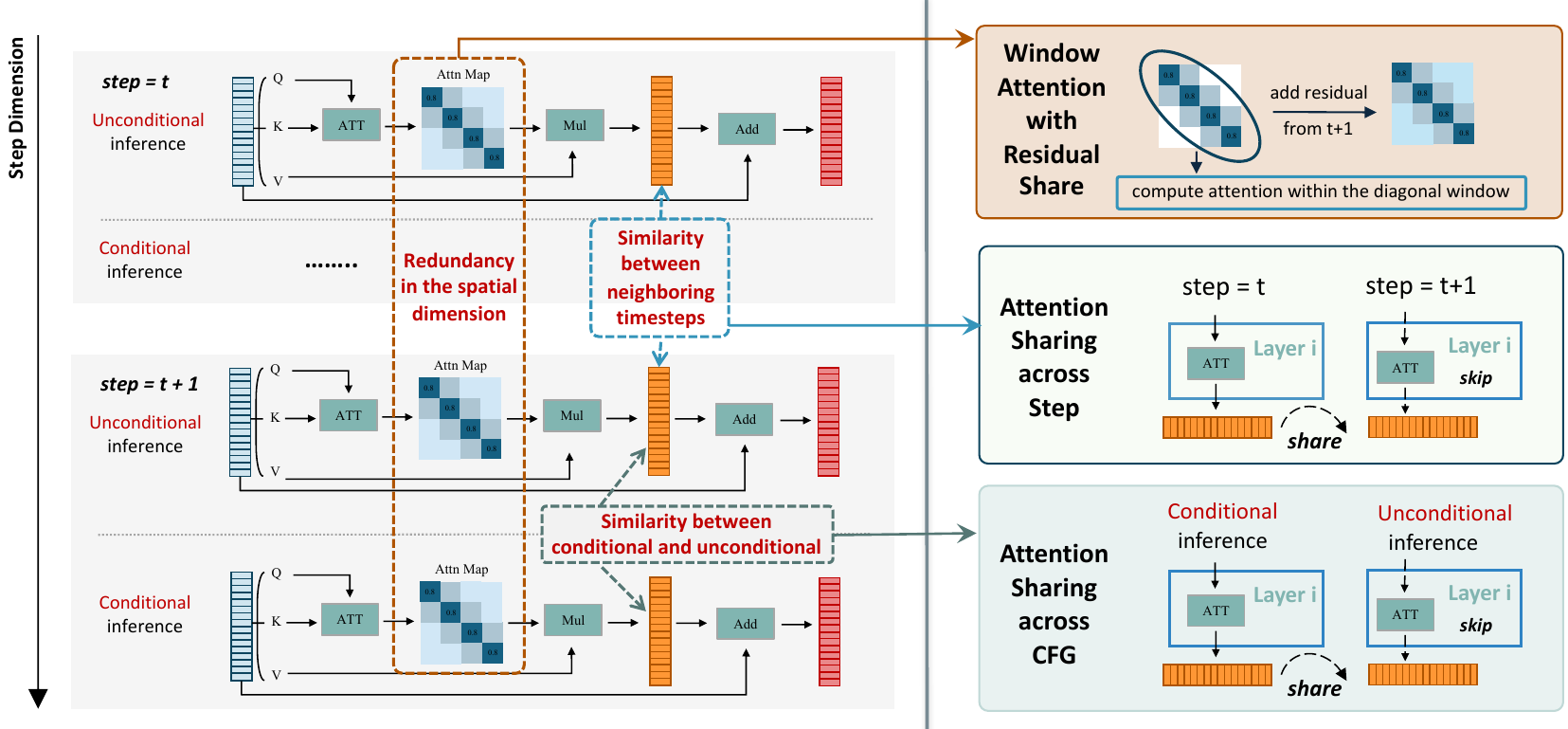}
    \caption{ \textbf{Types of redundancy and  corresponding compression techniques}. \textbf{Left:} Redundancy in the spatial dimension, denoising steps, and CFG. \textbf{Right:} Techniques implemented in \method{} to reduce redundancy for each type. 
\method{} employs window attention to minimize attention redundancy, while maintaining performance using residuals. Additionally, attention outputs are shared both step-wise and CFG-wise to reduce redundancy.}
    \label{fig:overview}
\end{figure}

% \begin{table}
% % \small
% \centering
% \caption{\textbf{tmp.}  .}
% \label{tab:decoder-latency}
% \begin{tabular}{cccccc} 
% \hline\hline
%  & Compression Dimension            & Memory Overhead & Speed Up~ & Speed Up Boundary    \\ 
% \midrule
% Attention             & Transformer & Low             & Low       & \textasciitilde{}10  \\ 
% \midrule
% CFG                   & Diffusion  & Low             & High      & 2                    \\ 
% \midrule
% Step              & Diffusion  & High            & High      & \textasciitilde{}10  \\
% \bottomrule
% \end{tabular}
% \end{table}
% 在这里我们在训练有素的扩散模型的推理步骤中揭示了diffusion model计算过程中的redundancy。从Diffusion model的推理过程中，我们定位了三种不同的冗余，如图~\ref{}所示：1.attention计算维度的冗余：我们发现attention map呈现明显的局部性； attnmap 的值主要集中在对角线的一个窗口之内； 2.classifier-free-guidance维度的冗余： 我们发现conditinal inference 和unconditional inference的attention map具有高度相似性；3.step维度的冗余：同一layer在相邻step输出之间的强相似性促使我们在相邻step之间共享输出。针对这三个维度的冗余性，我们分别提出了三种压缩方法，如图~\ref{}所示：1.attention计算维度的压缩：attention map的窗口局部性，启发我们只计算这个窗口内的attention值，而压缩其余部分的计算。我们使用Residual Window Attention 代替full attention，以压缩不必要的计算；2.CFG维度的压缩：conditinal attention map和unconditional attention map的高度相似性启发我们在conditinal和unconditional 推理。conditinal和unconditional之间的相似性，促使我们在两次推理之间共享attn map，以节省获取注意力图所需要的计算；3.step维度的压缩：相邻step输出之间的强相似性促使我们在相邻step之间共享输出，跳过attention layer的计算以加速模型。
In this section, we demonstrate the redundancy in the inference process of diffusion models with transformers. In the denoising process, we identify three types of redundancy, as shown in Figure~\ref{fig:overview}:
(1)\textit{ Redundancy in the spatial dimension.} 
% The attention map shows significant locality; its values are mainly concentrated within a diagonal window.
(2)\textit{ Similarity between the neighboring steps in attention outputs.} 
% The outputs of the same layer at neighboring steps show significant similarity.
(3)\textit{ Similarity between the conditional and unconditional inference in attention outputs.} 
% The outputs of the attention layer show clear similarities between conditional and unconditional inference.
% , prompting us to share outputs between adjacent steps.
To address these redundancies, we propose three compression techniques, as shown in Figure~\ref{fig:overview}: 
(1) In Sec. \ref{wars_section}, we introduce \textbf{\wars} to reduce spatial redundancy.
(2) In Sec. \ref{ass_section}, we introduce \textbf{\ass} to exploit step-wise similarities, thereby enhancing model efficiency.
(3) In Sec. \ref{asc_section}, we introduce \textbf{\asc} to reduce redundancy by utilizing similarity between conditional and unconditional generation.
% (1) in sec. \ref{wars_section}, we introduce
% \textbf{ \wars:} to reduce the reduncy in the spatial dimension.
%  (2) in sec. \ref{ass_section}, we introduce
% \textbf{ \ass :} to leverage similarity us to skip the attention layer computations at some steps and speed up the model. 
%  (3) in sec. \ref{asc_section}, we introduce
% \textbf{\asc :}reduce redundant computation in conditional and unconditional generation.
% Note that due to the residual connection, these output share methods do not produce identical predictions.
In Sec. \ref{calibration}, we introduce a simple greedy method to decide the compression plan, i.e., select the appropriate compression technique for each layer and step. 
% Utilizing a greedy algorithm, this method selects compression methods according to their impact on the final output.
% 在在这里，我们在训练有素的扩散模型的推理步骤中揭示了交叉注意的作用和功能首先在第 4.1 节中凭经验证明对交叉注意力图收敛的观察，然后在第 4.2 节中对观察结果进行了系统的分析。。扩散过程存在冗余，这些冗余分布在不同的维度。在本工作中，我们定位了三种不同的冗余：1. 2. 3.。如图~\ref{}所示，窗口注意力（Window Attention）是 xxx。变压器层（Transformer Layer）在不同时间步的计算是冗余的。CFG 是 xxx。

% 我们提出了减少冗余和降低推理成本的方法。残差窗口注意力方法 xxx。

% \begin{document}

% \end{document}

% There is redundancy in Diffusion Process. The redundancy resides in different dimensions. In this work, we localize the three different redundancy 1. 2. 3. . 
% As shown in Figure~\ref{}. Window Attention is xxx. computation of Transformer Layer in different step is redundant. CFG is xxx. 

% And propose methods for reduce the redundancy and reduce the inference cost. Method residual window attention xxx.

\subsection{\wars{} (\warsshort{})}
% Window Attention with Residual Share}

\label{wars_section}

%在以ViT为backbone的diffusion model中许多attention head具有local性质。在超过半数的head的attention map我们发现了window pattern。这说明许多head的图像token与相邻的数个token具有较高的attention score。因此在inference过程中对于一部分transformer block的使用fix-size window attention代替full attention可以保留大部分attention matrix中数值较高的部分，且由于只计算与相邻token的attention，根据window size的大小我们可以线性的减少self attention的计算量。

%然而在implementation的过程中我们发现单纯使用window attention的情况下，需要设置一个较大的attention window size来保留attention数值较高的部分（figure x左图）。在这种情况下，由于window 较大，减少的计算量较少，应用window attention带来加速效果不明显。而强行减小window size以提高加速比则会导致模型的性能下降，使得模型生成的图片会与原始模型生成的图片有较大的差异。这表明需要应用一种方法来保留图像token的global dependencies

\begin{figure}[tb]
    \centering
    \includegraphics[width=\textwidth]{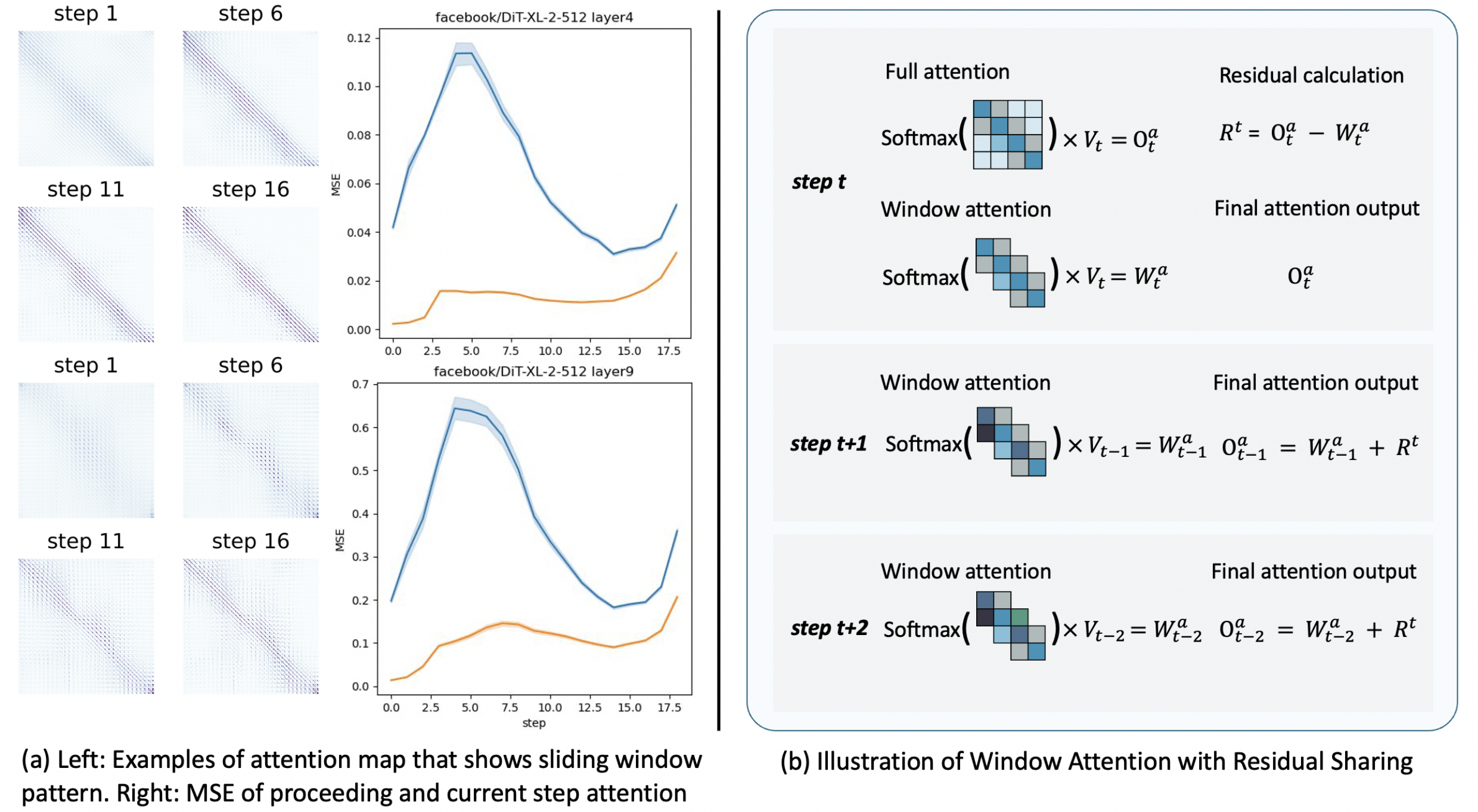}
    \caption{\textbf{\wars}. (a) \textbf{Left}: Example of the attention map showing the window pattern. \textbf{Right}: The MSE between the window attention outputs in the previous and current step (yellow line) versus the MSE between the output residuals of window and full attention in the previous and current step (blue line). The output residual exhibits minimal changes over the steps. 
    %The attention values within the window attention change significantly across steps, while the
    %Values within the sliding window change significantly across steps. \textbf{Right: MSE of proceeding and current step attention output: Residual vs. Window Attention} The residual values change minimally over time.
    (b) Computation of \wars{}. Window attention that illustrates significant changes is recalculated. Residuals that change minimally are cached and reused in subsequent steps.}
    \label{fig:residual_share}
\end{figure}

We can observe the spatial locality of attention in many transformer layers in pre-trained DiTs. As shown in Fig. \ref{fig:residual_share}(a), attention values concentrate within a window along the diagonal region of the attention matrix. Therefore, replacing full attention with fixed-size window attention for some layers can preserve most of the values in the attention matrix during inference. By computing attention values only within a specified window, the computation cost of attention can be largely reduced.

%The locality of attention is observed across transformer layers in well-trained diffusion transformer models. As shown in Fig. \ref{fig:residual_share}(a) attention values concentrate within a window along the diagonal region of the attention map. Therefore, replacing full attention with fixed-size window attention for some transformer blocks can retain most of the values in the attention matrix during inference. By computing attention values only within a specified window, the computation cost of attention can be linearly reduced based on the window size.
% However, in the implementation process, we found that some tokens still attend to a small set of spatial distant tokens. Directly discarding these dependencies by reducing the window size will negatively impact model performance and result in the inability to generate usable images. To mitigate the problem solely using window attention requires setting a relatively large attention window size to capture these dependencies. In such cases, the large window size results in a minimal reduction in computation cost, making acceleration difficult. 

However, some tokens still attend to a small set of spatial distant tokens. Discarding these dependencies negatively affects model performance. Mitigating this issue using only window attention necessitates a large window size to capture these dependencies. Consequently, this approach achieves minimal reduction in computational cost, thereby hindering acceleration efforts.

\textbf{Cache and Reuse the Residual for Window Attention.}
% 为了解决这一问题，我们提出Cache Residual for Window Attention，一种简单高效的方法与window attention相结合来在维持模型性能的同时最大程度的减少attention计算量。这一方法的基本思想是对于表现出window pattern的layer，我们利用attention score在相邻step 具有相似性这一特点，复用除window部分的数值。具体来说，就是对数值较大的window部分在每个step重新计算，而对于window之外的部分只在初始的step中计算一次而在并在后续相邻的step中共享这些数值。在一些特定的时间步上我们同时计算window attention和full attention得到O_{t} 和O_{t}^{window}。将两者相减得到残差R_{t}。这个残差将被保留并传递到后续的时间步中。在后续连续的几个step中模型仅需要计算window attention。后续step的output O_{t-1}将是window attention的output加上之前保留的残差结果。通过在模型中应用这一方法，我们可以缩减window size的大小。
To address the aforementioned issue, we investigate the information loss caused by using window attention. 
As shown in Fig. \ref{fig:residual_share}(a), the residual between the outputs of full and window attention exhibits a small variation across steps, unlike the output of window attention. This observation motivates us to cache the residual of window attention and full attention in one step and reuse it in subsequent steps. 

Fig.~\ref{fig:residual_share}(b) illustrates the computation of \warsshort{}: at each step, for each window attention layer, we compute the window attention and add a residual cached from the previous step to the output. 
We denote the set of steps that share the residual value \(\mathbf{R}_{r}\) as \(\mathbf{K}\), the full attention at step \( r \) as \(\mathbf{O}_{r}\), the window attention at step \(k\) as \(\mathbf{W}_{k}\).
%The process of \wars{} can be divided into two parts: computing the residual and applying window attention with the residual. 
For the first step in the set \(r = \min \mathbf{(K)}\), the computation of \warsshort{} goes as follows:
%\begin{gather}
\begin{equation}
\begin{aligned}
    \mathbf{O}_{r} &= \mathrm{Attention}(\mathbf{Q}_r,\mathbf{K}_r,\mathbf{V}_r),  \\
    \mathbf{W}_{r} &= \mathrm{WindowAttention}(\mathbf{Q}_r,\mathbf{K}_r,\mathbf{V}_r),  \\
    \mathbf{R}_{r} &= \mathbf{O}_{r} - \mathbf{W}_{r}.
    \end{aligned}
\end{equation}
For a subsequent step in the set \(k \in \mathbf{K}\), the computation of \warsshort{} goes as follows:
\begin{equation}
\begin{aligned}
    \mathbf{W}_{k} &= \mathrm{WindowAttention}(\mathbf{Q}_k,\mathbf{K}_k,\mathbf{V}_k), \\
    \mathbf{O}_{k} &= \mathbf{W}_{k} + \mathbf{R}_r.
\end{aligned}
\end{equation}

%Incorporating Cache Residual for Window Attention facilitates the reduction of window size while upholding generation performance.

% In subsequent steps, the model only needs to compute window attention. The output \( O_{t-1} \) in subsequent steps is the output of window attention plus the previously retained residual result: \\
% \begin{gather}
%     O^a_{t-1} = W^a_{t-1} + R_t 
%     % O^f_{t-1} = FFN(O^a_{t-1})
% \end{gather}

% \subsubsection{Kernel Launch}

% \subsection{Heads Attention Share}
% %从similarity 1Calibration、2group内共享attention map、3attention map与各个K获得结果，简单提一下kernel launch
% \subsubsection{Calibration}
% %最大最远采样、每一层分多少组的算法

% \subsubsection{Efficient Attention Share by QK Merge}
% %attention map计算方法，merge q,k的方法，simple average qk or L2-Norm as regularization、

\subsection{\ass{} (\assshort{})}
\label{ass_section}
\begin{figure}[h]
    \centering
    \includegraphics[width=\textwidth]{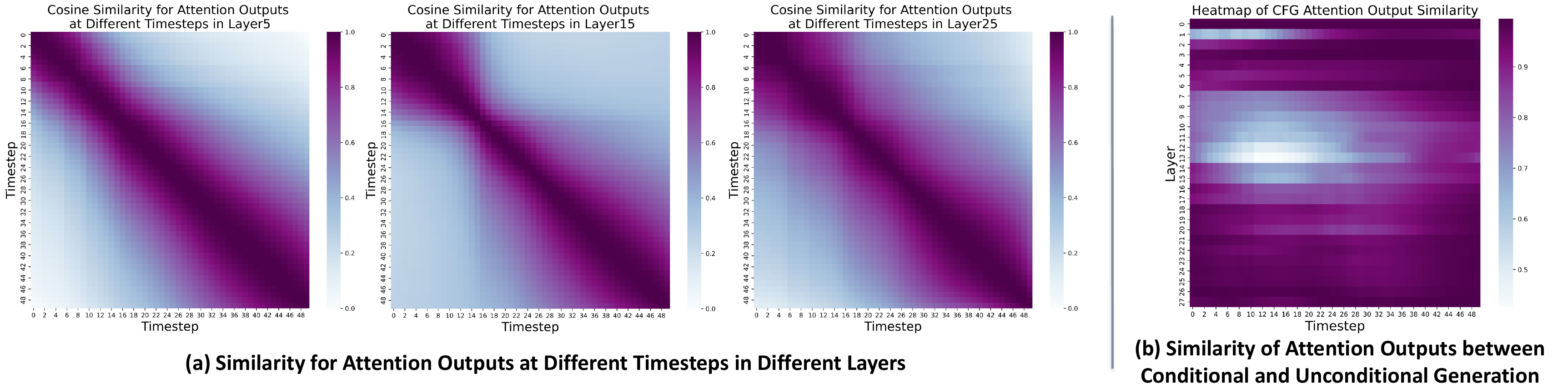}
    \caption{ \textbf{Similarity of Attention Outputs Across Step and CFG Dimensions in DiT.}
(a) Similarity of attention outputs across step dimension in different layers. (b) Similarity between conditional and unconditional attention outputs in various layers at different steps}
    \label{fig:output_share}
\end{figure}

The sequential nature of the denoising process in diffusion models is a major bottleneck for inference speed.
Here, we compare the attention outputs at different steps during the denoising process. We find that for some layers, the attention outputs at certain steps show significant similarity to those of adjacent steps. Fig. ~\ref{fig:output_share}(a) presents the cosine similarity between the attention outputs at different steps. We can draw two primary observations: (1) There is a noticeable temporal similarity between the attention outputs; (2) This similarity varies across steps and layers.
% The inherent sequential nature of the denoising process in diffusion models is a major bottleneck for inference speed.
% Here, we compare the outputs of the attention layer at different steps during the denoising process. We find that some steps' outputs show significant similarity to adjacent steps. Fig. ~\ref{fig:output_share}(a) shows the similarity between features of the attention layer outputs at different steps during DiT denoising. Through observation, we get two primary insights: (1) There is a noticeable temporal feature similarity between adjacent steps in the denoising process; (2) This similarity varies across different steps and layers.
% Early and late time steps show higher similarity, while the mid time steps show lower similarity.
% The similarity between attention outputs indicates redundant calculations that can be optimized.  We think that allocating significant computational resources to regenerate similar attention outputs leads to high costs with minimal benefits. Inspired by DeepCache~\citep{ma2023deepcache} and considering the variation in similarity across steps and layers, we propose an effective, training-free technique to reuse attention outputs.

To exploit this similarity to reduce the computational cost, we propose the \assshort{} technique. Specifically, for a set of steps with their attention outputs similar to each other, we cache the earliest step's attention output \(\mathbf{O}\) and reuse it, thereby skipping the computation at the subsequent steps. 

% \subsubsection{Kernel Launch}
\subsection{\asc{} (\ascshort{})}
\label{asc_section}
% 扩散模型去噪过程固有的顺序性是推理速度的主要瓶颈之一。在本节中，我们回顾了去噪过程中生成的特征，发现去噪过程中的相邻步骤在高级特征中表现出显着的时间相似性。在图\todo{}中，我们提供了与这一观察结果相关的经验证据。实验阐明了去噪过程中相邻步骤之间存在明显的时间特征相似性，表明连续步骤之间的变化通常很小，这表明某些高级特征以渐进的速度变化。这种现象可以在大量成熟的模型中观察到，如DiT、Pixart-alpha。基于这些观察结果，我们的认为利用这一相似性来加速去噪过程。我们的分析表明，计算通常会导致特征与上一步的特征非常相似，从而突出了优化的冗余计算的存在。我们认为，分配重要的计算资源来重新生成这些类似的特征图构成了一个效率低下的问题。需要大量的计算开销，但产生的边际效益很低。
% classifier free guidance(CFG)被广泛使用，如DALL·E 2和Imagen，显著提高样本生成质量。在推理时，CFG分别进行了条件生成和无条件生成两次前向过程。在本节中，我们对比了条件生成和无条件生成过程，试图揭示可以优化注意力图生成过程以提高推理效率。观察：在去噪过程中时间步后期，条件生成和无条件生成在注意力图上表现出显着的相似性。在图\todo{}中，我们提供了与这一观察结果相关的经验证据。实验阐明了两个主要见解：1）条件生成和无条件生成在注意力图上表现显著相似性，表明时间步后期，条件导致的图像变化通常很小； 2）无论我们使用的扩散模型如何，在后期时间步，对于每个层，条件生成和无条件生成的注意力图都显示出高度相似性（&ssim;0.95）。这种现象可以在大量成熟的模型中观察到，如DiT、Pixart-alpha，这表明在去噪过程后期图片轮廓清晰的情况下，条件 guidance信息已经逐渐蕴含在噪声图像中，从而突出了重复推理的冗余计算的存在。基于这些观察结果，我们的目标是利用这一相似性来加速去噪过程。我们的分析表明，计算通常会导致特征与上一步的特征非常相似，从而突出了CFG两次前向过程的冗余计算的存在。我们认为，分配重要的计算资源来重新生成这些类似的注意力图构成了一个效率低下的问题。需要大量的计算开销，但产生的边际效益很低，但它为扩散模型速度的效率提高提出了一个潜在的领域。
% 不同的step、instance有一致性 -> offline Calibration
% simple分组的结果（maybe simple Kmeans
%Classifier-free guidance (CFG) is widely used, such as in DALL·E 2~\citep{ramesh2022hierarchical} and Imagen~\citep{saharia2022photorealistic}. 
Classifier-free guidance (CFG) is widely used for conditional generation~\citep{ho2022classifier,ramesh2022hierarchical,saharia2022photorealistic}. 
%significantly improving generation quality. 
In each step of the inference process for conditional generation, CFG performs two neural network inferences: one with the conditional input and one without. This doubles the computational cost compared with unconditional generation.
As shown in Figure \ref{fig:output_share}(b), for many layers and steps, the similarity between the attention outputs in the conditional and unconditional neural network evaluations is high. %(\(\approx 0.95\)). 

Based on this observation, we propose the \ascshort{} technique that reuses the attention output from the conditional neural network evaluation in the unconditional neural network evaluation. 
%We cache the conditionally generated attention output, $\mathbf{O_{cond}}$. During unconditional generation, we reuse the cached $\mathbf{O_{cond}}$ as the unconditionally generated attention output, $\mathbf{O_{ucond}}$ and skip the computation of $\mathbf{O_{ucond}}$. This technique can be formulated as:
% \begin{gather}
%     \mathbf{O_{ucond} = O_{cond}}
% \end{gather}
% The similarity between conditional and unconditional generation changes with time steps. We developed an algorithm to automatically decide whether to share results between conditional and unconditional generation.
 % Our analysis indicates that the attention maps of conditional and unconditional generation are very similar, highlighting the redundancy in CFG's two forward passes.
% \subsection{Steps Attention Share}
% This phenomenon can be observed in many mature models, such as DiT and Pixart-alpha.?
% Residual Attention Output Cache (ResA Cache)

%TODO
\subsection{Method for Deciding the Compression Plan}
\label{calibration}
% 上述提到的Cache Residual for Window Attention，transformer output share，CFG attention score share三种方法均能在保持模型性能的情况下加速推理过程。这三种方法适用于不同的layer不同step，为了将这三种方法结合起来，确定每个layer在相应step上使用的方法，我们提出了一个基于calibration的方案。我们使用一小部分的标签/text prompt作为calibration dataset，生成相应的图片。在这个过程中对每个layer各step的hidden state进行标注。具体流程如下，当step为0时计算full attention，对于后续的step，尝试应用window attention + cache residual，cfg 和 output share三种方法，测试应用方法生成的图像与原始生成图像的structural similarity。保留达到ssim阈值并加速效果最好的方案。若所有方案均未达到设定的ssim阈值，说明hidden states变化较大，无法共享。这种情况下重新计算full attention。

% These aforementioned techniques, including \wars, \ass, and \asc, can accelerate the inference process while maintaining model performance. These techniques are applicable to different layers and steps. However, as shown in Fig.\ref{fig:output_share}, significant variations in similarity are observed across different layers and steps. Therefore, it is crucial to develop an algorithm to automatically decide which techniques should be applied at each layer and step. We developed a greedy algorithm to decide which compression technique should be applied by measuring its impact on the final result. 

The aforementioned techniques, including \warsshort{}, \assshort{}, and \ascshort{}, can effectively cut down the computational cost while maintaining the performance. As shown in Fig.~\ref{fig:residual_share} and Fig.~\ref{fig:output_share}, different layers have different redundancies in different time steps. Therefore, it is crucial to properly decide the compression plan, i.e., which techniques should be applied for each layer at each step.
%These techniques are applicable to different layers and steps. However, as shown in Fig.\ref{fig:output_share}, significant variations in similarity are observed across different layers and steps. Therefore, it is crucial to develop an algorithm to automatically decide which techniques should be applied at each layer and step. We developed a greedy algorithm to decide which compression technique should be applied by measuring its impact on the final result. 

We develop a simple greedy method to select the appropriate strategy (a combination of techniques) from a strategy list $\mathcal{S}=$[\assshort{}, \warsshort{} + \ascshort{}, \warsshort{}, \ascshort{}] for each step and each layer. 
As shown in Alg.~\ref{alg:greedy}, we determine the strategies step by step and layer by layer. 
For each step and transformer layer, we apply each of the four compression strategies and calculate the loss between the model outputs with and without compression for the current step, $L(O,O')$. Then, we select the strategy with the highest computation reduction ratio with loss below a threshold $\frac{i}{|M|}\delta$, where $i$ is the layer index and $|M|$ is the number of layers in the model. If none of the four strategies meet the threshold, we do not apply compression for this layer at that step.

%The algorithm evaluate each combination of compression techniques applied to each layer at each step by measuring the MSE between the final outputs with and without compression. The combination that reaches the MSE threshold $\delta$ and provides the best acceleration is selected. For a continuous sequence of steps employing either the \wars{} or the \ass{} technique, we aggregate them into a set \( \mathbf{K} \).
%If none of the technique combinations reach the threshold $\delta$, it indicates that each compression technique significantly degrades model performance. Therefore, no compression technique should be applied, and full attention with CFG should be applied. Our proposed greedy-based algorithm allows us to select the most appropriate compression techniques for each layer at each step, optimizing network efficiency while maintaining generation performance. The cost of the algorithm is low. The cost is \#Techniques$\times$\#Layers times of the cost for an original image generation.

\begin{algorithm}
    \SetKwInOut{Input}{Input}
    \SetKwInOut{Output}{Output}

    % \underline{Greedy Compression Selection} $(M,T,l,\delta)$\;
    \Input{Transformer Model $M$, Total Step $T$, Compression Strategy List $\mathcal{S}$, Threshold $\delta$}
    \Output{dictionary \texttt{dict} that stores selected compression techniques}
    Initialize \texttt{dict}
    
    \For{step $t$ in $T$}
    {
        $O$ $\leftarrow$ \text{compute the output of the uncompressed $M$}
        
        \For{transformer layer $i$ in $M$}
        {
            \For{$m$ $\in$ $S$ \text{order by ascending compression ratio}}
            {
            \text{compress layer $i$ in step $t$ using compression strategy $m$}
                
            $O'$ $\leftarrow$ \text{compute the output of $M$}
                
                \If{$\text{L}(O, O') < \frac{i}{|M|}\delta$}
                {
                \text{update $m$ as the selected strategy of layer $i$ and step $t$ in \texttt{dict}}
                
                \text{break}
                }
            }
        }
    }
    \Return{} \texttt{dict}
    \caption{Method for Deciding the Compression Plan}
    \label{alg:greedy}
\end{algorithm}
% If none of the methods reach the set SSIM threshold, indicating significant variations in hidden states that prevent sharing, recalibrate by recomputing full attention. This calibration-based approach allows us to dynamically select the most suitable method for each layer at each step, optimizing both performance and efficiency.

% Our calibration algorithm is as follows:

% \begin{enumerate}[itemsep=0.1\baselineskip]
%     \item Sort compression method combinations by computational cost in descending order.
%     \item Iterate over each step of each attention layer, \( attn_t \).
%     \item Apply each sorted compression method combination \( c \) to \( attn_t \).
%     \item Compare the SSIM of the network's output before and after compression.
%     \item If the SSIM is greater than the threshold, replace \( attn_t \) with \( c \). Otherwise, return to step 3.
% \end{enumerate}
\section{Experiments}
%为了证明我们方法的有效性与预训练的diffusio transformer 的类型无关，我们在三个常用的diffusion transformers上评估我们的方法：DiT, Pixart-Alpha和Open-Sora。除此之外，为了证明我们的方法与快速采样方法兼容，我们在Dit的20步DPM-Solver、Pixart-Alpha的25步DPM-Solver, Open-Sora的250步和50步PLMS上构建了我们的方法.  We select MS-COCO 2017 as image generation dataset and xxx as video generation dataset. For other datasets, we generate 50k images to assess the generation quality. We follow previous works [10, 55, 66] to employ the evaluation metrics including FID and IS. we assess the latency per sample on a platform equipped with a Nvidia A100 40G.

\subsection{Settings}

We evaluate \method{} on three commonly used diffusion transformers: DiT~\citep{peebles2023scalable} and Pixart-Sigma~\citep{chen2024pixart} for image generation tasks, and Open-Sora~\citep{Open-Sora} for video generation tasks. To demonstrate compatibility with fast sampling methods, we build our method upon 50-step DPM-Solver for DiT and Pixart-Sigma, and 200-step IDDPM~\citep{nichol2021improved} for Open-Sora. %For better comparison with the raw model, 

For calculating quality metrics, we use ImageNet as the evaluation dataset for DiT and MS-COCO as the evaluation dataset for PixArt-Sigma. MS-COCO 2014 caption is used as text prompt for Pixart-Sigma models' image generation. 
% We use xxx as the video generation dataset.
To evaluate generation quality, we generate 50k images for DiT models and 30k images for PixArt-Sigma models. 
% and for the xxx dataset, we generated xx videos to evaluate generation quality. 
Following previous studies, we employ FID~\citep{heusel2017gans}, IS~\citep{salimans2016improved} and CLIP score~\citep{hessel2021clipscore} as the evaluation metrics. We measure the latency per sample on a single Nvidia A100 GPU.

%We denote Window Attention with Residual Share as WARS, Attention Sharing across Steps as ASS, and Attention Sharing across CFG as ASC.
We use mean relative absolute error for $L(O,O')$ and experiment with and different thresholds $\delta$ at intervals of 0.025. %to decide the selected compression techniques.
We denote these threshold settings as D1 ($\delta$=0.025), D2 ($\delta$=0.05), ..., D6 ($\delta$=0.15), respectively. We set the window size of \warsshort{} to 1/8 of the token size.

\subsection{Results on Image Generation}
\textbf{Results of Evaluation Metrics and Attention FLOPs.} \method{} was applied to the pre-trained DiT-XL-2-512, PixArt-Sigma-1024, and PixArt-Sigma-2K models. Table \ref{tab:imgen_score} displays the evaluation results of these models. For the DiT-XL-2-512 and PixArt-Sigma-1024 models, configurations D1, D2, and D3 nearly matched the performance of the original models in terms of IS and FID metrics. %However, models D4, D5, and D6, despite achieving higher compression rates, showed a slight decline in IS and CLIP scores. 
% The IS and CLIP scores for settings D4, D5, and D6 decrease slightly as a trade-off for achieving higher compression rates. The attention computation of DiT-XL-2-512 was reduced to 84.8\%, 68.8\%, 58.9\%, 48.8\%, 40.9\%, and 33.6\% from D1 to D6, respectively; for PixArt-Sigma-1024, it was reduced to 89.9\%, 73.5\%, 62.6\%, 51.9\%, 44.2\%, and 37.4\%.
% The PixArt-Sigma-2K model maintained performance close to the original up to D5, with a noticeable performance decline only at D6, where attention computation was reduced to 80.5\%, 59.7\%, 45.8\%, 35.7\%, 28.9\%, and 24.3\%. 
Comparison of compression effects and evaluation metrics between the three models reveals that as image resolution increases, \method{} not only achieves greater compression but also better preserves the generative performance of the models.

\begin{figure}[t]
    \centering
    \includegraphics[width=0.98\textwidth]{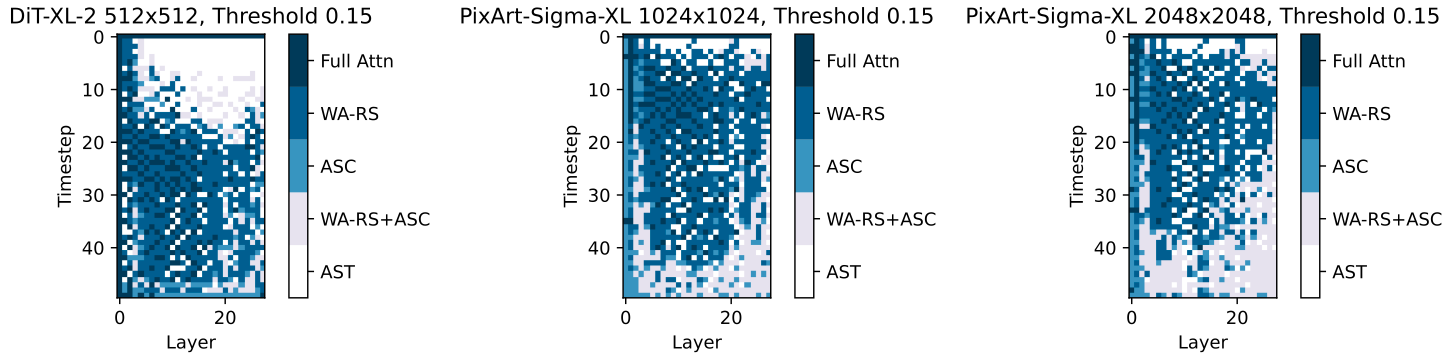}
    \caption{Compression plan for DiT-XL-512, PixArt-Sigma-XL-1024 and PixArt-Sigma-XL-2K at D6 with the number of DPM-Solver steps set to 50.}
    \label{fig:plan_comparison}
\end{figure}

\textbf{Compression Plan after Search.}
Figure~\ref{fig:plan_comparison} illustrates the compression plan under the D6 setting. For the DiT model, AST and ASC are utilized in the early timesteps, with full attention primarily appearing in the initial attention layers. In contrast, the PixArt-Sigma model employs AST sporadically in the first two layers and in the middle attention layers during the intermediate timesteps, while the combination of WA-RS and ASC is notably predominant in the final timesteps. This variability in the distribution of different types of redundancy across models highlights the absence of a universal compression strategy, underscoring the necessity for tailored plan searches. Additional compression plans for other settings are provided in the Appendix.

\begin{figure}[t]
    \centering
    \includegraphics[width=0.9\textwidth]{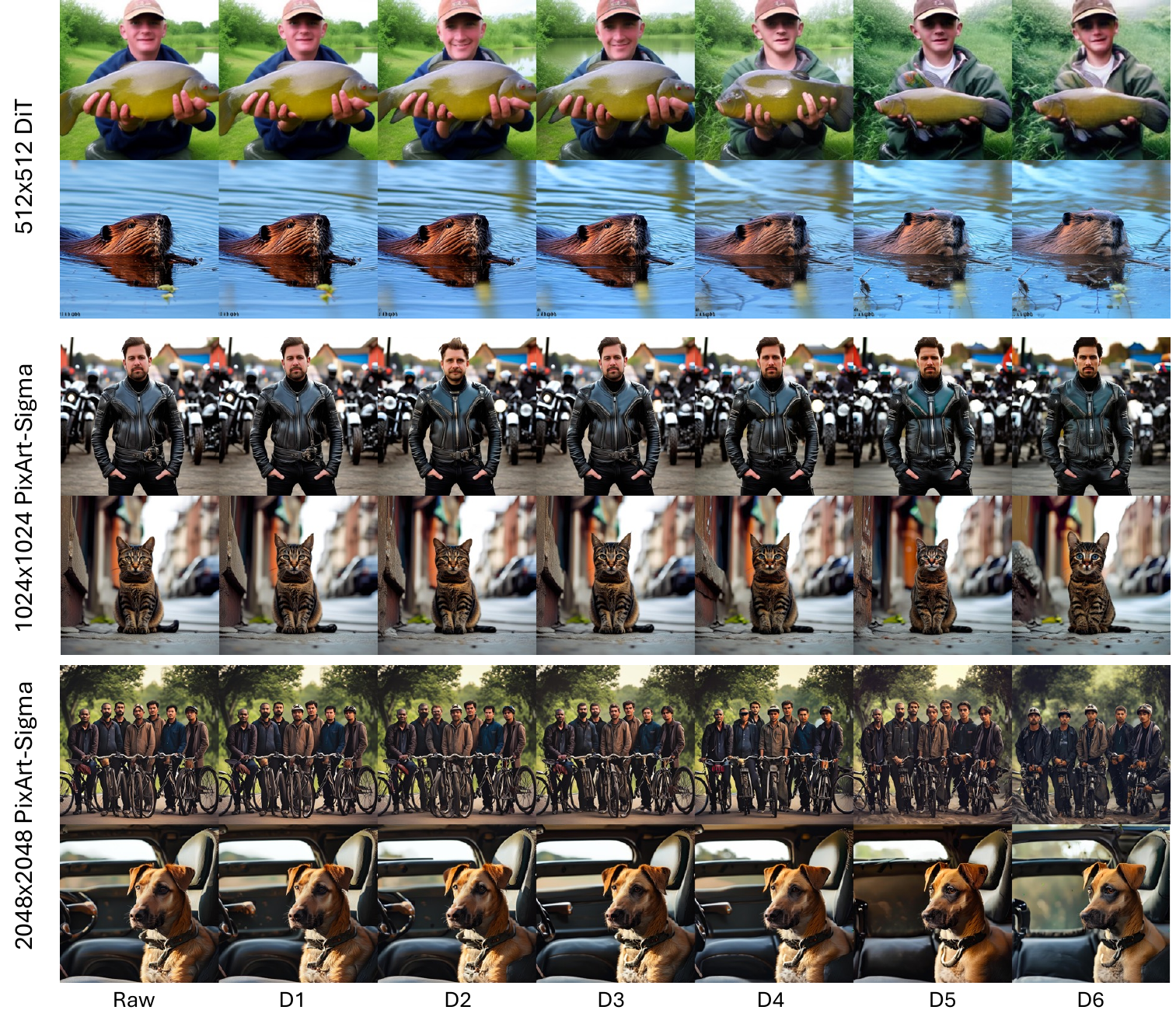}
    \caption{Image generation samples at various image resolutions under various compression ratios.}
    \label{fig:imgen_samples}
\end{figure}

\textbf{Visualization of \method{}'s Generation Results.} 
Figure \ref{fig:imgen_samples} presents image generation samples from \method{}. The D1, D2, and D3 configurations of the DiT-XL-2-512 and PixArt-Sigma-1024 models demonstrate visual generation quality comparable to the original models. In contrast, D4, D5, and D6 achieve greater compression, exhibiting slight variations in detail while still producing acceptable-quality images. The PixArt-Sigma-2K model maintains image quality similar to the original up to D4, with D5 and D6 also generating high-quality outputs. This suggests that our compression method effectively preserves generation quality, even when reducing attention computation by over 50\% and compressing to 33\% at higher resolutions.

\begin{table}[t]
\centering
\caption{Image generation performance of \method{} at various image resolutions under various compression ratios. The FID, IS, and CLIP results are marked in different makers. The "Attn FLOPs" represents the fraction of computation in the multi-head attention module compared to the raw model.} %Performance of different resolutions of image generation under different compression ratios. }
\label{tab:imgen_score}
\resizebox{1\linewidth}{!}{
\begin{tabular}{@{}c|ccc|cccc|cccc@{}}
\toprule
Model & \multicolumn{3}{c|}{DiT-XL-2 512x512} & \multicolumn{4}{c|}{PixArt-Sigma-XL 1024x1024} & \multicolumn{4}{c}{PixArt-Sigma-XL 2048x2048} \\ \midrule
Score & IS     & FID   & \begin{tabular}[c]{@{}c@{}}Attn\\FLOPs\end{tabular} & IS    & FID   & CLIP  & \begin{tabular}[c]{@{}c@{}}Attn\\FLOPs\end{tabular} & IS    & FID   & CLIP  & \begin{tabular}[c]{@{}c@{}}Attn\\FLOPs\end{tabular} \\  \midrule
Raw   & 408.16     & 25.43    & 100\%         & 24.33    & 55.65    & 31.27    & 100\%         & 23.67    & 51.89    & 31.47    & 100\%        \\
D1    & 412.24     & 25.32    & 85\%          & 24.27    & 55.73    & 31.27    & 90\%          & 23.28    & 52.34    & 31.46    & 81\%         \\
D2    & 412.18     & 24.67    & 69\%          & 24.25    & 55.69    & 31.26    & 74\%          & 22.90    & 53.01    & 31.32    & 60\%         \\
D3    & 411.74     & 23.76    & 59\%          & 24.16    & 55.61    & 31.25    & 63\%          & 22.96    & 52.54    & 31.36    & 46\%         \\
D4    & 391.80     & 21.52    & 49\%          & 24.07    & 55.32    & 31.24    & 52\%          & 22.95    & 51.74    & 31.39    & 36\%         \\
D5    & 370.07     & 19.32    & 41\%          & 24.17    & 54.54    & 31.22    & 44\%          & 22.82    & 51.21    & 31.34    & 29\%         \\
D6    & 352.20     & 16.80    & 34\%          & 23.94    & 52.73    & 31.18    & 37\%          & 22.38    & 49.34    & 31.28    & 24\%         \\ \bottomrule
\end{tabular}
}
\end{table}

% \begin{figure}[t]
%     \centering
%     \includegraphics[width=0.96\textwidth]{images/dit_result.png}
%     \caption{ \textbf{The 512x512 image generation with DiT-XL-2-512. Left: Inception Score (IS) and FID value. Right: Image comparing under different compression ratio.} }
%     \label{fig:dit_result}
% \end{figure}

% \begin{figure}[t]
%     \centering
%     \includegraphics[width=0.96\textwidth]{images/pixart1k_result.png}
%     \caption{ \textbf{The 1024x1024 image generation with PixArt-Sigma-XL-2-1024-MS. Left: Inception Score (IS), FID value, and CLIP score. Right: Image comparing under different compression ratio.} }
%     \label{fig:pixart1k_result}
% \end{figure}

% \begin{figure}[t]
%     \centering
%     \includegraphics[width=0.96\textwidth]{images/pixart2k_result.png}
%     \caption{ \textbf{The 2048x2048 image generation with PixArt-Sigma-XL-2-2K-MS. Left: Inception Score (IS), FID value, and CLIP score. Right: Image comparing under different compression ratios.} }
%     \label{fig:pixart2k_result}
% \end{figure}

\subsection{Results on Video Generation}

\begin{figure}[t]
    \centering
    \includegraphics[width=0.98\textwidth]{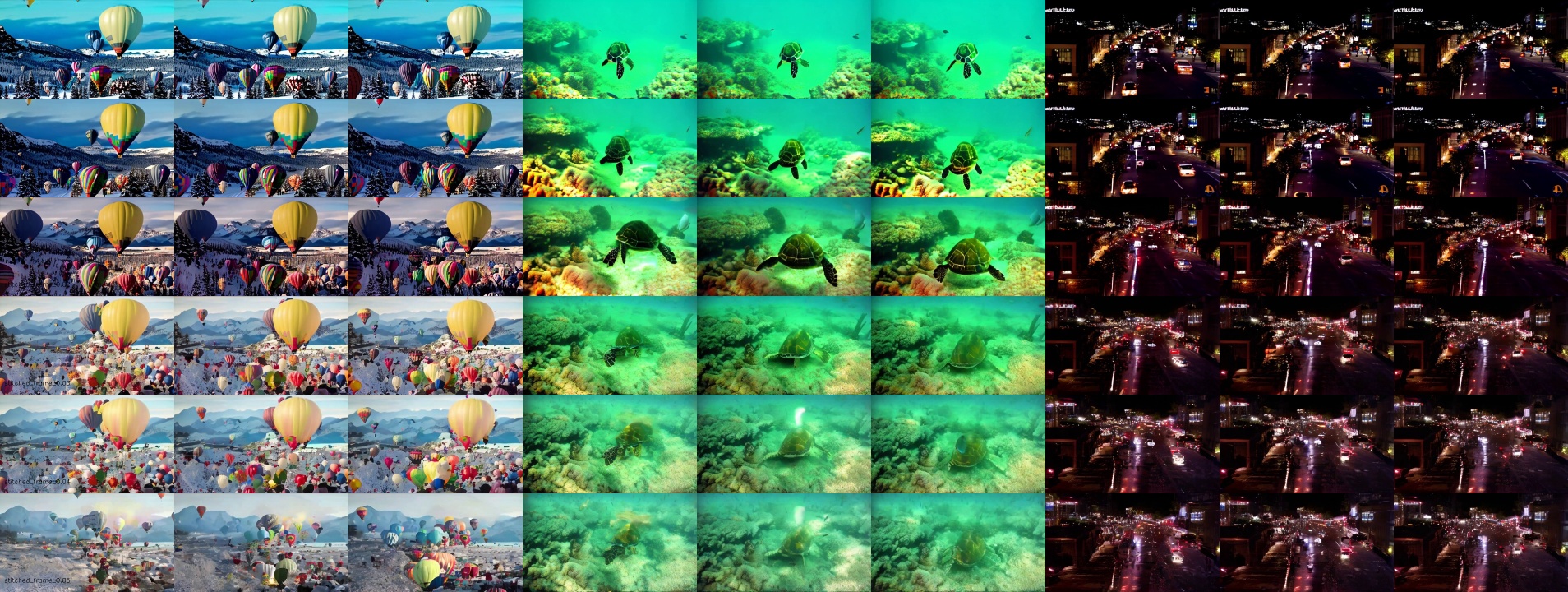}
    % \caption{Generated videos (OpenSora V1.1 16 frames 240p), the top is raw video, and the middle is D3 configuration and the bottom is D5 configuration.}   
    \caption{Comparison of video generation using OpenSora V1.1 at 240p resolution with 16 frames.}
    \label{fig:opensora_result}
\end{figure}

We apply DitFastAttn on OpenSora for video generation with five thresholds from 0.01 to 0.05. The results are shown in Figure~\ref{fig:opensora_result}. 
Specifically, the reduction in computation for configurations labeled D1 through D6 are as follows: 7.63\%, 19.50\%, 30.16\%, 37.66\%, 40.52\%, respectively. For an extended analysis, additional results are provided in the Appendix.
% The attention computation cost on 240p video generation given configuration of D1 to D6 are 88.19\%, 73.55\%, 59.98\%, 53.46\%, 50.90\%, and 48.06\% respectively. More results are in Appendix.

\begin{table}[tb]
\centering
\caption{FLOPs fraction and latency fraction of DitFastAttn in Diffusion Transformers comparing with original attention. The latency is evaluated on the Nvidia A100 GPU.}
\label{table:flops_speedup}
\resizebox{0.9\linewidth}{!}{
\begin{tabular}{@{}ccccccc@{}}
\toprule
Model                                & Seqlen                 & Metric  & \ascshort{} & \warsshort{}   & \warsshort{}+\ascshort{} & \assshort{} \\ \midrule
\multirow{2}{*}{DiT-XL-2 512x512}        & \multirow{2}{*}{1024}  & Attn FLOPs   & 50\% & 77\% & 38\%    & 0\% \\
                                     &                        & Attn Latency & 59\% & 85\% & 51\%    & 4\% \\ \midrule
\multirow{2}{*}{PixArt-Sigma-XL 1024x1024} & \multirow{2}{*}{4096}  & Attn FLOPs   & 50\% & 51\% & 26\%    & 0\% \\
                                     &                        & Attn Latency & 54\% & 54\% & 31\%    & 3\% \\ \midrule
\multirow{2}{*}{PixArt-Sigma-XL 2048x2048}   & \multirow{2}{*}{16384} & Attn FLOPs   & 50\% & 33\% & 16\%    & 0\% \\
                                     &                        & Attn Latency & 52\% & 35\% & 19\%    & 1\% \\ \bottomrule
\end{tabular}
}
\end{table}

\subsection{\#FLOPs Reduction and Speedup}
% Tab.\ref{table:flops_speedup} show FLOPs fraction and latency fraction of DitFastAttn in Diffusion Transformers comparing with original attention. ASC 可以将attention计算量降为50\%, latency的下降随着分辨率的增大而有轻微增大。 ASC随着分辨率的增大可以将attention计算量降为从69\% 到 23\%, latency的下降随着分辨率的增大从93\%下降到70\%。WA和ASC方法是互相正交的，两者可以同时使用，两者同时使用不会带来额外的开销，Flops和Latency收到两个方法的效果综合同比降低。AST直接跳过了Attention的计算，latency的下降随着分辨率的增大而有轻微增大。
%Fig. \ref{fig:imgen_samples} 展现了raw和D1到D6的，\method{}对overall image generation随着计算量减少的latency的减少，和总体attention 随着计算量减少的latency减少。 in DiT-XL,在D1，D2，D3虽然FLOPs下降，但image generation和Attention的overall latency反而有轻微的上升。we note \method{}在推理时没有引入额外的开始，latency的上升是由于我们的kernel没有well implement,相比FlashAttention-2的实现性能有所下降.D3之后latency相比raw有所下降，D6取得了最低的latency,overall generation latency约为raw的94.27\%, 总体Attention latency为raw的82.32\%. PixArt-Sigma-1024随着计算量下降，latency持续下降。PixArt-Sigma-2K取得了最好的性能，在D6，overall generation latency 为raw的76.32%； overall attention latency为raw的64。27%。随着分辨率的增大\method{}在总体attention和image generation的latency上都能取得更好的性能。值得note的是，如果我们进一步优化我们的kernel implementation，我们可以取得更好的latency reduction。
\textbf{Compression Results of \method{} on Various Sequence Lengths.} We implement DiTFastAttn based on FlashAttention-2~\citep{dao2023flashattention}. Table \ref{table:flops_speedup} shows the FLOPs fraction and latency fraction of DiTFastAttn in Diffusion Transformers compared with original attention mechanisms. The \ascshort{} technique reduces attention computation by 50\%, with latency reduction slightly increasing as resolution increases. As resolution increases, \warsshort{} can reduce attention computation from 77\% to 33\%, and latency reduction ranges from 85\% to 35\%. The \warsshort{} and \ascshort{} techniques are orthogonal; they can be used simultaneously without additional overhead.
% resulting in a proportional decrease in both FLOPs and latency due to the combined effect of the two methods. 

\begin{figure}[t]
    \centering
    \includegraphics[width=0.95\textwidth]{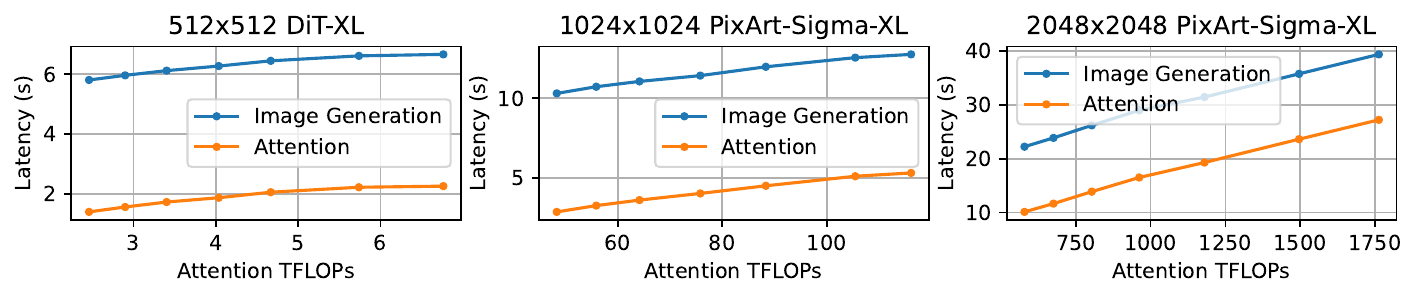}
    \caption{Latency of different resolutions of image generation under different compression ratios. DiT runs with a batch size of 8, while PixArt-Sigma models with a batch size of 1. The blue line delineates the latency for end-to-end image generation, whereas the orange line represents the latency of multi-head attention module.}
    \label{fig:latency_plot}
\end{figure}

\textbf{Overall Latency of \method{}.} Figure \ref{fig:latency_plot} shows the latency for image generation and attention as computation decreases when \method{} is applied. \method{} achieves end-to-end latency reduction for all three model at all compression ratio settings. PixArt-Sigma-2K shows the best performance, with overall generation latency at D6 being 56\% of raw and overall attention latency at 37\%. The result indicates that as resolution increases, \method{} achieves better performance in reducing latency for both overall attention and image generation.

\begin{figure}[t]
    \centering
    \includegraphics[width=0.95\textwidth]{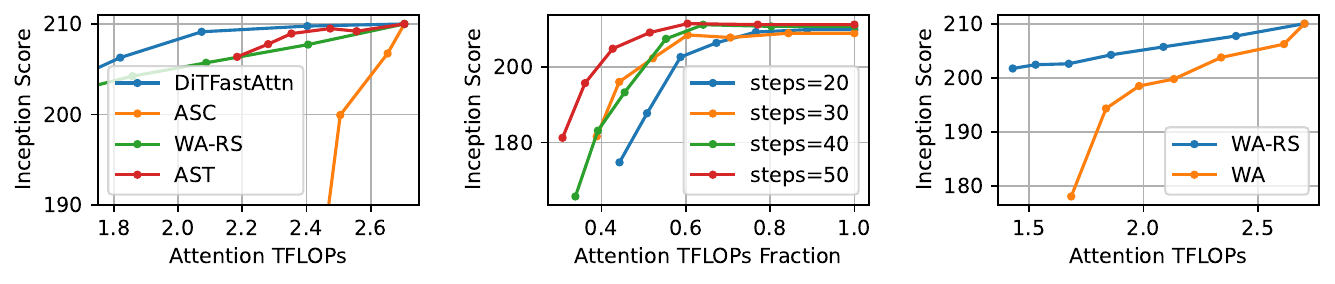}
    \caption{Ablation study on DiT-XL-2-512. 
    % The left is the experiment on methods. The middle is the experiment on different timesteps. The right is the experiment on the residual share. WA means the window attention without the residual share (RS). 
    Examination of methodological impact (Left), timesteps variability (Middle), and residual sharing (Right). "WA" denotes Window Attention without the Residual Share (RS).}
    \label{fig:ablation}
\end{figure}

\subsection{Ablation Study}
% \begin{figure}[t]
%     \centering
%     \includegraphics[width=0.65\textwidth]{images/ablation.pdf}
%     \caption{Ablation Study on DiT-XL.}
%     \label{fig:ablation}
% \end{figure}
\textbf{\method{} outperforms single methods}. 
As shown on the left of Fig. \ref{fig:ablation}, \method{} maintains higher quality metrics compared to individual techniques with the same computation budget. Among single techniques, AST shows the best generation quality. However, beyond 2.2 FLOPs, further compression using AST significantly degrades the outputs, causing the search algorithm to terminate. \method{} supports further compression while maintaining better quality.

\textbf{Higher steps improve \method{}'s performance}. As shown on the middle of Fig. \ref{fig:ablation}, we compared the performance of \method{} at different steps. It is evident that as the step increases, \method{} can compress more computation while maintaining quality.

\textbf{The residual caching technique is essential in maintaining the performance}. As shown on the right of Fig. \ref{fig:ablation}, \wars{} maintains better generative performance than \wa{} at the same compression ratio. Without residuals, window attention results in a significant performance drop.

% Number of 

% \subsubsection{Attention Categorization}

% \begin{table}[]
% \begin{tabular}{llll}
% Model                    & FID     & IS       & SSIM \\
% pretrained               & 31.8689 & 214.5040 & 1.00 \\
% static (threshold = 0.9) & 31.7827 & 213.8928 & 0.91 \\
% static (threshold = 0.8) & 31.5919 & 214.0936 & 0.84 \\
% static (threshold = 0.7) & 31.2192 & 212.3567 & 0.78
% \end{tabular}
% \end{table}

% \subsubsection{Timesteps Attention Share}

% \begin{table}[]
% \begin{tabular}{llll}
% Model                     & FID     & IS       & SSIM \\
% pretrained                & 32.0805 & 214.0324 & 1.00 \\
% dynamic (threshold = 0.9) & 32.0718 & 212.3210 & 0.88 \\
% dynamic (threshold = 0.8) & 32.0759 & 211.7968 & 0.81 \\
% dynamic (threshold = 0.7) & 32.1048 & 211.9636 & 0.75
% \end{tabular}
% \end{table}

\section{Conclusion}
% 在本文中，我们引入了一种新的post-training方法 DiTFastAttention 来加速扩散模型。我们定位了三种不同的冗余并propose 对应的 compression methods：1.attention计算维度的冗余; 我们提出了Window Attention with Residual Share来减少冗余的同时保持性能 2.Step维度的冗余；我们考虑了attention输出在step-wise相似性，DiTFastAttention提出了step Attention Output Share 来跳过冗余的Step计算.3.CFG维度的冗余；我们考虑了attention输出 CFG-wise的相似性；并通过在条件推理和非条件推理间共享输出，减少冗余的计算。通过 DiTFastAttention 的采用，实现了计算速度的显着加速。对几个数据集和扩散模型的实证评估表明，DiTFastAttention 超越了其他attention计算压缩方法。
% This is conclusion.
In this paper, we introduce a novel post-training compression method, DiTFastAttention, to accelerate diffusion models. We identify three types of redundancy : 
(1) Redundancy in the spatial dimension. 
% The attention map shows significant locality; its values are mainly concentrated within a diagonal window.
(2) Similarity between the neighboring steps in attention outputs.
% The outputs of the same layer at neighboring steps show significant similarity.
(3) Similarity between the conditional and unconditional inference in attention outputs. 
And we propose corresponding compression techniques: (1) \wars, (2) \ass, (3) \asc.
The experiments show that DiTFastAttention significantly reduces the cost of attention and accelerates computation speeds.

\textbf{Limitations.} First, our method is a post-training compression technique and therefore cannot take advantage of training to avoid the performance drop. Second, our method mainly focuses on inference acceleration instead of VRAM reduction. When AST is applied, the attention hidden states from previous timestep will be stored and will bring extra VRAM usage. Third, our simple compression method may not find the optimal compression plan. Fourth, our method only reduces the cost of attention module. % In future work, we plan to explore training-aware compression methods. We also aim to extend our approach to other modules. Additionally, further kernel-level optimizations may unlock even greater speedups for the proposed compression techniques.

\section*{Acknowledgements}
This work was supported by National Natural Science Foundation of China (No. 62325405, 62104128,
U19B2019, U21B2031, 61832007, 62204164), Tsinghua EE Xilinx AI Research Fund, and Beijing
National Research Center for Information Science and Technology (BNRist). %We thank for all the
%support from Infinigence-AI.

\clearpage

{\small
\bibliographystyle{iclrbib}
\bibliography{reference}
}

\clearpage
\newpage
\appendix
\section{Appendix}
\label{sec:appendix}

\subsection{Societal Impacts} 
DiTFastAttention can enable more efficient deployment of diffusion transformer models, which have shown remarkable image and video generation capabilities. On the positive side, accelerating these models could democratize access to powerful generative AI by reducing computational requirements, and allowing broader adoption for creative and educational applications. However, there are also potential negative societal impacts that must be carefully considered. Highly realistic synthetic media could be exploited to create deepfakes for misinformation, fraud, or non-consensual editing. There are also potential privacy risks if generative models can reconstruct personal information from data. While DiTFastAttention does not inherently increase or reduce these risks compared to the original models, widening access makes misuse by malicious actors more likely. Therefore, safeguards against misuse and ethical guidelines for the responsible release of compressed models may be needed as this technology develops.

% \subsection{Release Safeguards} 
% While DiTFastAttention itself does not release new pre-trained models, the compression capabilities it provides could enable easier sharing and deployment of powerful diffusion models that have risks of misuse. To mitigate risks of misuse, we have implemented access control. Users must agree to terms prohibiting unethical applications.

% \subsection{Codes} 
% Our codebase is anonymously available in the Supplementary Materials. 
% For implementation details, we have them added to readme documentation in the codebase. 

\subsection{Details of the Method for Deciding the Compression Plan}

This algorithm has a computation complexity of $\mathcal{O}(|\mathcal{S}| \times |T| \times |M|^2 \times s^2)$, where $|\mathcal{S}|$ is the number of compression strategies (4 in our case), $|T|$ is the number of denoising steps, $|M|$ is the number of transformer layers, and $s$ is the sequence length. While the inference time for generating an image using the DiT has a computation complexity of $\mathcal{O}(|T| \times |M| \times s^2)$. Therefore, the greedy algorithm takes about $|\mathcal{S}|\times|M|$ of the image generation time. For example, the inference time for generating a 512$\times$512 image using DiT-XL-2-512 is approximately 2 seconds, so the greedy algorithm takes around 224 seconds (2s $\times$ 28 transformer layers $\times$ 4 method candidates) to decide the compression plan, which is a reasonable overhead compared to the overall inference time.

% The mean relative absolute error is calculated as $L(O,O')=\frac{1}{|O|_1}\sum_i \text{clip}(\frac{O_i-O'_i}{\max(O_i,O'_i)+\epsilon},0,10)$, where $|O|_1$ is the number of elements in $O$, $\epsilon$ is a small number (1e-6 in our experiments) and clip is a function to clip the value into [0,10] to avoid numeric unstability.

The mean relative absolute error is a metric used to evaluate the performance of a model by measuring the average relative deviation between the outputs $O'$ and the raw outputs $O$. It is calculated as follows:
\begin{equation*}
L(O, O') = \frac{1}{|O|_1} \sum_{i} \text{clip}\left(\frac{|O_i - O'_i|}{\max(|O_i|, |O'_i|) + \epsilon}, 0, 10\right)
\end{equation*}
In this equation, $|O|_1$ represents the number of elements in the raw output vector $O$. The summation iterates over each element $i$ in the vectors $O$ and $O'$. For each element, the absolute difference between the raw output $O_i$ and the output $O'_i$ is calculated. This difference is then divided by the maximum value between $|O_i|$ and $|O'_i|$, which serves as a normalization factor to make the error relative to the magnitude of the output values.
To avoid numerical instability in cases where both $O_i$ and $O'_i$ are very small or zero, a small positive constant $\epsilon$ (set to $10^{-6}$ in our experiments) is added to the denominator. The $\text{clip}$ function ensures that the resulting ratio is clipped to the range [0, 10], preventing extreme values from dominating the overall error.
The clipped ratios are summed and then divided by the total number of elements $|O|_1$ to obtain the mean relative absolute error. This metric provides a normalized measure of the average relative deviation between the predicted and raw outputs, with values ranging from 0 to 10 (maximum allowed error).

\subsection{Results for Video Generation}

\begin{figure}[t]
    \centering
    \includegraphics[width=0.8\textwidth]{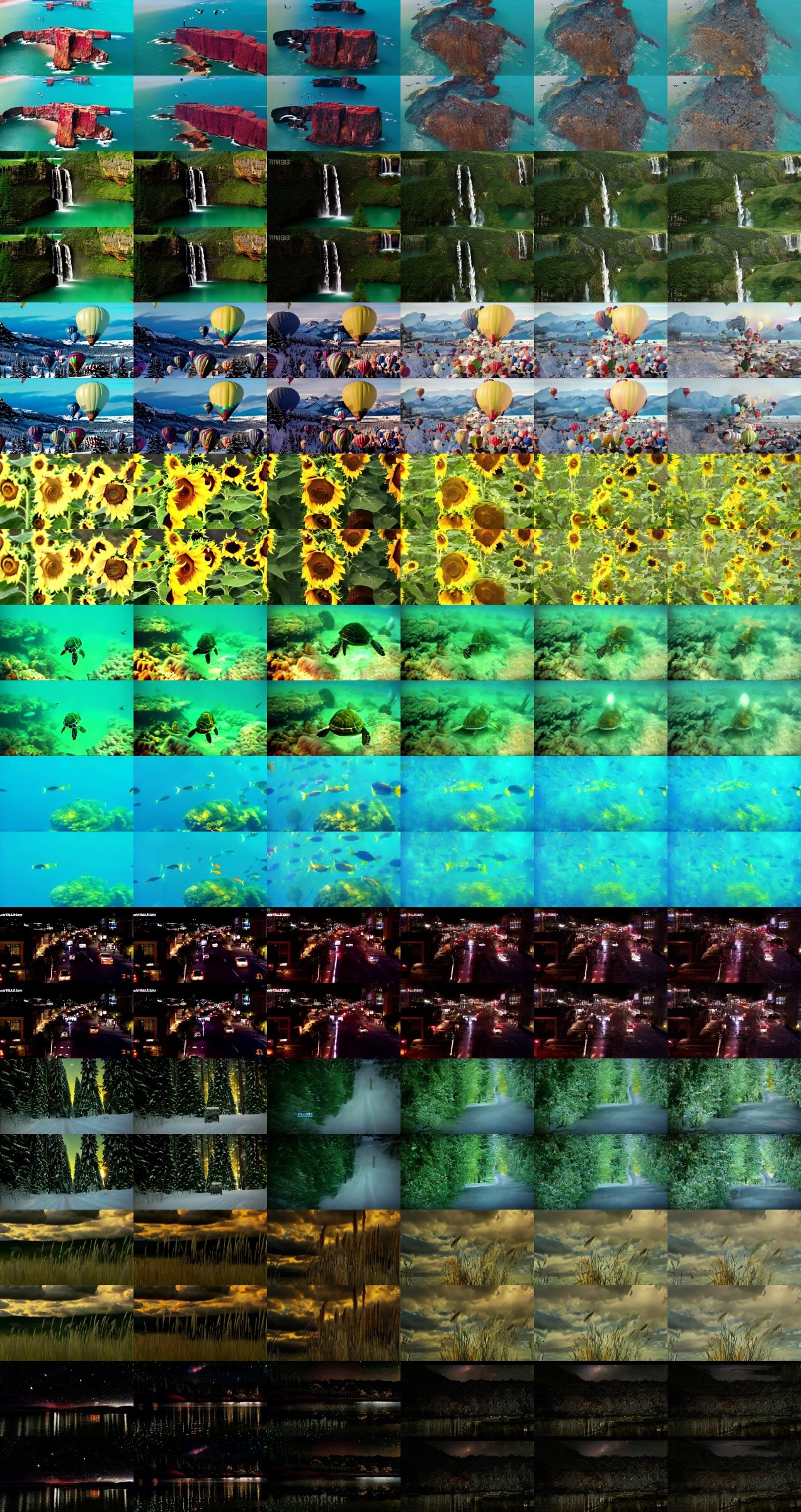}
    % \caption{Generated videos (OpenSora V1.1 16 frames 240p), the top is raw video, and the middle is D3 configuration and the bottom is D5 configuration.}
    \caption{Comparison of video generation using OpenSora V1.1 at 240p resolution with 16 frames. The left column displays the original video, and the right columns illustrate the outputs from the D1 
 to D6 configuration.}
    \label{fig:opensora_result_appendix}
\end{figure}

As shown in Figure~\ref{fig:opensora_result_appendix}, the subjective evaluation of DiTFastAttn's application to video generation tasks revealed a significant performance distinction across configurations D1 through D6. 
Configurations D1 to D4 demonstrated effective performance, balancing computational efficiency with the retention of visual quality in generated videos. 
The subjective assessment indicated that videos generated under these configurations were smooth, with natural transitions between frames and preserved details that are critical for video quality. The maintenance of these qualities suggests that the model was able to effectively leverage the redundancies identified and apply the compression techniques without noticeable loss to the viewer.

% In contrast, configurations D5 and D6, which applied more aggressive compression, The generated image can be different from the original one in color and position of objects.
% However, the generated video is also smooth and can represent the prompt.

In contrast, configurations D5 and D6, which applied more aggressive compression techniques, resulted in a noticeable deviation from the original video characteristics. The generated videos under D5 and D6 were still smooth and coherent, allowing them to represent the intended narrative or prompt with reasonable accuracy. This suggests that while aggressive compression can compromise certain aspects of video quality, it can still be effective in scenarios where computational resources are limited and a high level of detail is not paramount.

The subjective results underscore the importance of finding an optimal balance between computational efficiency and generation quality when applying DiTFastAttn to video generation tasks. While configurations D1 to D4 offer a promising trade-off, the deviation observed in D5 and D6 highlights the need for careful consideration of the compression parameters.
For practical deployment, it is crucial to select a DiTFastAttn configuration that aligns with the specific requirements of the application in terms of both performance and output quality.
% Further refinement of the compression methods may expand the range of effective compression ratios, making DiTFastAttn an even more versatile tool for accelerating video generation tasks.

\subsection{Latency values in different settings}

% facebook/DiT-XL-2-512 (batch size=8) Image generation inference latencies are [Raw=2.88s, D1=2.9s, D2=2.96s, D3=2.9s, D4=2.82s, D5=2.79s, D6=2.71s].
% Attention inference latencies are [Raw=0.91s, D1=0.93s, D2=0.95s, D3=0.92s, D4=0.83s, D5=0.8s, D6=0.75s].
% PixArt-alpha/PixArt-Sigma-XL-1024-MS (batch size=1) Image generation inference latencies are [Raw=2.44s, D1=2.44s, D2=2.45s, D3=2.42s, D4=2.32s, D5=2.25s, D6=2.18s].
% Attention inference latencies are [Raw=0.94s, D1=0.94s, D2=0.93s, D3=0.91s, D4=0.81s, D5=0.74s, D6=0.68s].
% PixArt-alpha/PixArt-Sigma-XL-2K-MS (batch size=1) Image generation inference latencies are [Raw=16.88s, D1=16.49s, D2=16.23s, D3=15.13s, D4=13.94s, D5=13.19s, D6=12.88s].
% Attention inference latencies are [Raw=11.23s, D1=10.92s, D2=10.57s, D3=9.5s, D4=8.36s, D5=7.58s, D6=7.22s].
% 在这里我们展示了不同setting下原始和使用DiTFastAttn的模型的延迟和视觉效果数值。值得注意的是原始的pixart-sigma就IS较小而FID较大，我们认为高分辨率以及选用的训练数据导致模型生成的图像分布和mscoco这类真实图像有较大区别。
\begin{table}[htbp]

\centering
\caption{Latency, FID and IS upon 50-step DPM-Solver. Attn Latency means the latency of self-attention. DiT-XL-2 runs with a batch size of 8. 50000 images used to generate FID and IS score}
\label{tab:result_table_dit}
\vspace{5px}

\begin{tabular}{lcccccc}
\toprule
Model & Resolution & Config & Latency (s) & Attn Latency (s) & FID & IS \\
\midrule
\multirow{7}{*}{DiT-XL-2} & \multirow{7}{*}{512x512} & Raw & 6.66 & 2.26 & 25.43 & 408.16 \\
 & & D1 & 6.61 & 2.22 & 25.32 & 412.24 \\
 & & D2 & 6.45 & 2.05 & 24.67 & 412.18 \\
 & & D3 & 2.89 & 0.91 & 23.76 & 411.74 \\
 & & D4 & 2.78 & 0.83 & 21.52 & 391.80 \\
 & & D5 & 2.77 & 0.80 & 19.32 & 370.07 \\
 & & D6 & 2.66 & 0.71 & 16.80 & 352.20 \\
\bottomrule
\end{tabular}%

\end{table}

\begin{table}[htbp]
\centering
\caption{Latency, FID and IS upon 50-step DPM-Solver. Attn Latency means the latency of self-attention. Models run with a batch size of 1. 30000 images used to generate FID, IS, and CLIP score}
\label{tab:result_table_pixart_sigma}

\begin{tabular}{ccccccc}
\toprule
Model                                                                                & Config & Latency (s) & Attn Latency (s) & FID   & IS    & CLIP  \\ \midrule
\multirow{7}{*}{\begin{tabular}[c]{@{}c@{}}PixArt-Sigma-XL\\ 1024x1024\end{tabular}} & Raw    & 12.76       & 5.30                                                   & 24.33 & 55.65 & 31.27 \\
                                                                                     & D1     & 12.55       & 5.10                                                   & 24.27 & 55.73 & 31.27 \\
                                                                                     & D2     & 11.98       & 4.49                                                   & 24.25 & 55.69 & 31.26 \\
                                                                                     & D3     & 11.42       & 4.01                                                   & 24.16 & 55.61 & 31.25 \\
                                                                                     & D4     & 11.06       & 3.60                                                   & 24.07 & 55.32 & 31.24 \\
                                                                                     & D5     & 10.73       & 3.25                                                   & 24.17 & 54.54 & 31.22 \\
                                                                                     & D6     & 10.31       & 2.85                                                   & 23.94 & 52.74 & 31.18 \\ \midrule
\multirow{7}{*}{\begin{tabular}[c]{@{}c@{}}PixArt-Sigma-XL\\ 2048x2048\end{tabular}} & Raw    & 39.86       & 27.57                                                  & 23.67 & 51.89 & 31.47 \\
                                                                                     & D1     & 35.75       & 23.62                                                  & 23.28 & 52.34 & 31.46 \\
                                                                                     & D2     & 31.44       & 19.29                                                  & 22.90 & 53.01 & 31.32 \\
                                                                                     & D3     & 28.99       & 16.51                                                  & 22.96 & 52.54 & 31.36 \\
                                                                                     & D4     & 26.18       & 13.88                                                  & 22.95 & 51.74 & 31.39 \\
                                                                                     & D5     & 23.86       & 11.66                                                  & 22.82 & 51.22 & 31.34 \\
                                                                                     & D6     & 22.27       & 10.13                                                  & 22.38 & 49.34 & 31.28 \\ \bottomrule
\end{tabular}

\end{table}

\begin{table}[htbp]
\centering
\caption{Latency, FID and IS under DiT paper experiment setting (250-step IDDPM solver, cfg scale = 1.5). Attn Latency means the latency of self-attention. DiT-XL-2 runs with a batch size of 12}
\label{tab:latency_fid_step50}

\resizebox{0.9\textwidth}{!}{%
\begin{tabular}{lcccccc}
\toprule
Model & Resolution & Config & Latency (s) & Attn Latency (s) & FID & IS \\
\midrule
\multirow{5}{*}{DiT-XL-2} & \multirow{5}{*}{512x512} & Raw & 32.62 & 11.40 & 3.16 & 219.97 \\
 & & D1 & 31.53 & 10.21 & 3.09 & 218.20 \\
 & & D2 & 29.35 & 8.09 & 3.10 & 210.36 \\
 & & D3 & 27.80 & 6.56 & 3.54 & 196.05 \\
 & & D4 & 26.96 & 5.77 & 4.52 & 180.34 \\
\bottomrule
\end{tabular}%
}
\end{table}

\begin{figure}[t]
    \centering
    \includegraphics[width=0.98\textwidth]{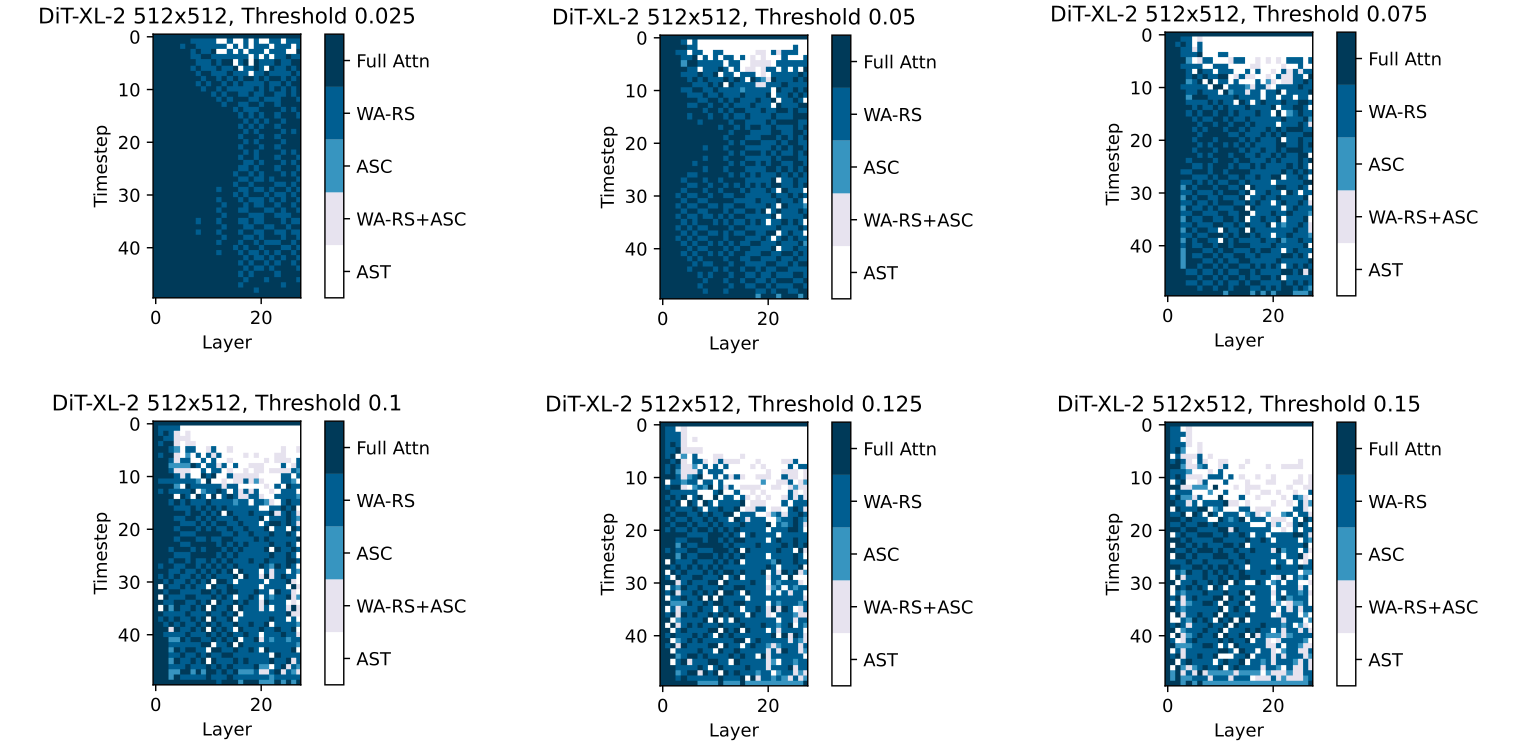}
    \caption{Compression plan for DiT-XL-2-512x512 at different thresholds with DPM solver step set to 50}
    \label{fig:dit_plan}
\end{figure}
\begin{figure}[t]
    \centering
    \includegraphics[width=0.98\textwidth]{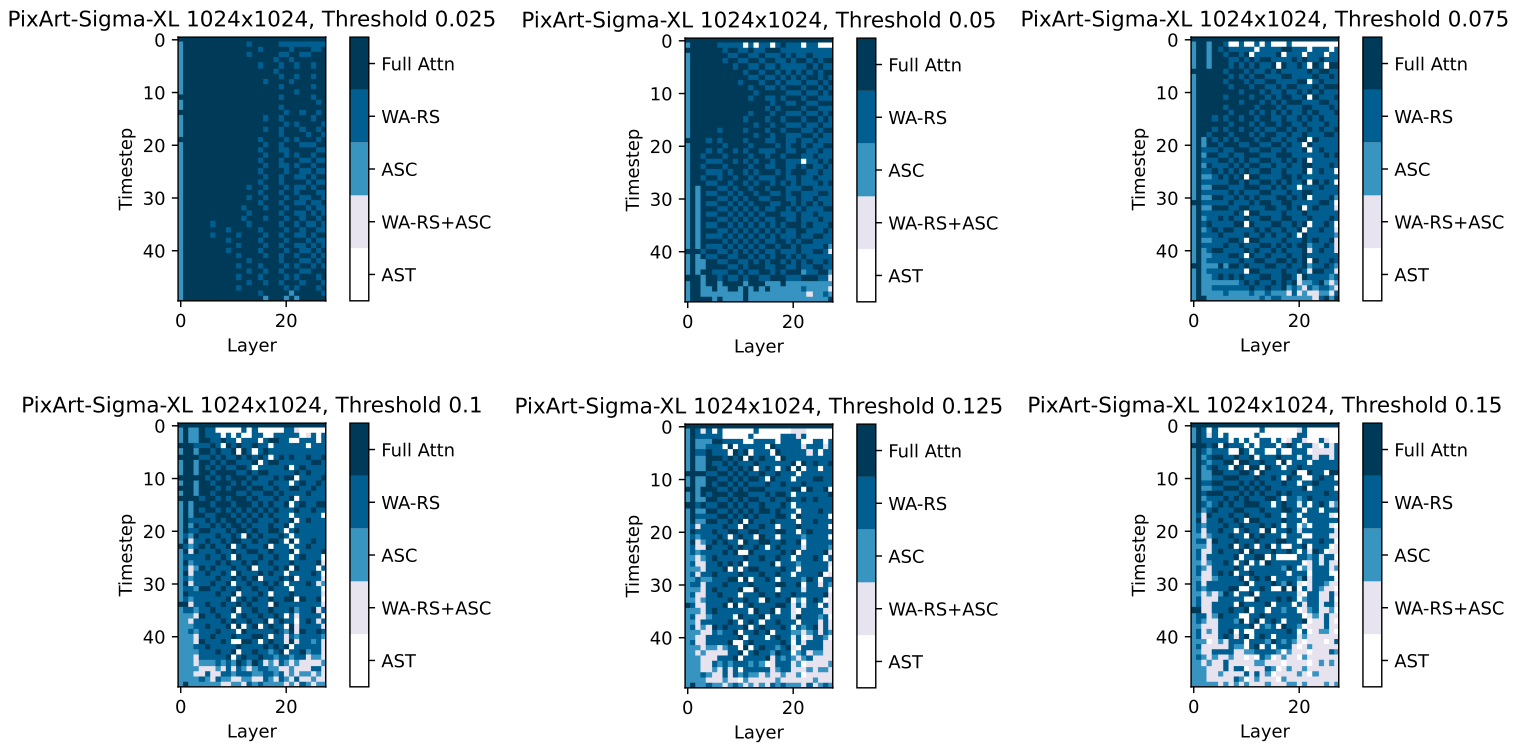}
    \caption{Compression plan for PixArt-Sigma-XL-1024x1024 at different thresholds with DPM solver step set to 50}
    \label{fig:pixart1k_plan}
\end{figure}
\begin{figure}[t]
    \centering
    \includegraphics[width=0.98\textwidth]{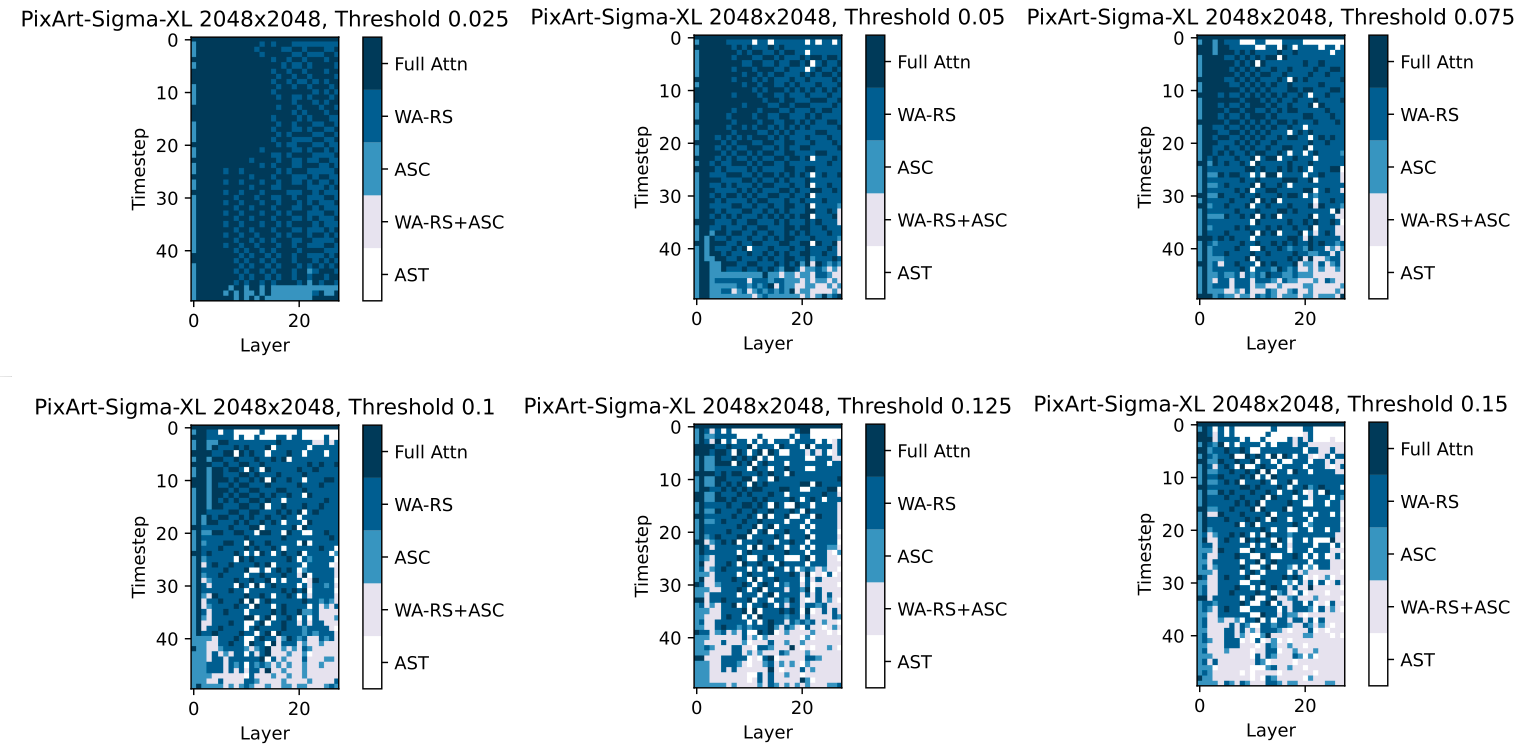}
    \caption{Compression plan for PixArt-Sigma-XL-2K at different thresholds with DPM solver step set to 50}
    \label{fig:pixart2k_plan}
\end{figure}

\subsection{Compression plan}
Figure~\ref{fig:dit_plan}, \ref{fig:pixart1k_plan}, \ref{fig:pixart2k_plan} display compression plan obtained after our greedy search method in different model settings as heatmaps. Each block stands for one layer at specific step. Both models exhibit three different kinds of redundancy, but the distribution of these redundancies across time steps and layers is quite different. The results indicate that there is no uniform compression plan for different DiT models and a search plan is essential in this case.

\subsection{Compression plan search time}
Table ~\ref{tab:search_time} show the plan search time in different configuration. Our greedy search method will try method that can achieve high compression ratio so plan search time is inversely proportional to threshold.

\begin{table}[htbp]
\caption{Compression plan search time for three models}
\label{tab:search_time}
\centering
\resizebox{0.7\textwidth}{!}{%
\begin{tabular}{lcccccc}
\toprule
Model & Resolution & Config & Plan Search Time \\
\midrule
\multirow{4}{*}{DiT-XL-2} & \multirow{4}{*}{512x512} & Raw & 04m39s \\
 & & D2 & 04m08s \\
 & & D4 & 03m49s \\
 & & D6 & 03m14s \\
\hline
\multirow{4}{*}{PixArt-Sigma-XL} & \multirow{4}{*}{1024x1024} & Raw & 22m02s\\
 & & D2 & 20m12s\\
 & & D4 & 17m50s\\
 & & D6 & 15m49s\\
\hline
\multirow{4}{*}{PixArt-Sigma-XL} & \multirow{4}{*}{2048x2048} & Raw & 1h50m13s\\
 & & D2 & 1h46m04s\\
 & & D4 & 1h22m53s\\
 & & D6 & 1h23m01s\\
\bottomrule
\end{tabular}%
}
\end{table}

\subsection{Metrics for compression plan search}

  When designing the compression plan, we have considered to use other metrics including LPIPS and SSIM, and finally chose the existing metric mainly because of the speed of computation. We tested different SSIM compression schemes and found that when SSIM is chosen as the metric, to ensure the quality of the images generated, the threshold should be set at a small value of about 1/10 of the existing metric. We checked LPIPS as an alternative metric by decoding the hidden states into RGB space then calculate LPIPS on RGB space. We found that LPIPS is very insensitive to value changes in our use cases. Only small value changes can be observed when switch diffrent methods and always suggest to use sharing across timestep when threshold is set to 0.005 or smaller. Addistionally, it takes a long time to compute LPIPS. In this way, we believe LPIPS is not a suitable metric for compression plan search. 

\subsection{Negative conditioning}

\begin{figure}[t]
    \centering
    \includegraphics[width=0.60\textwidth]{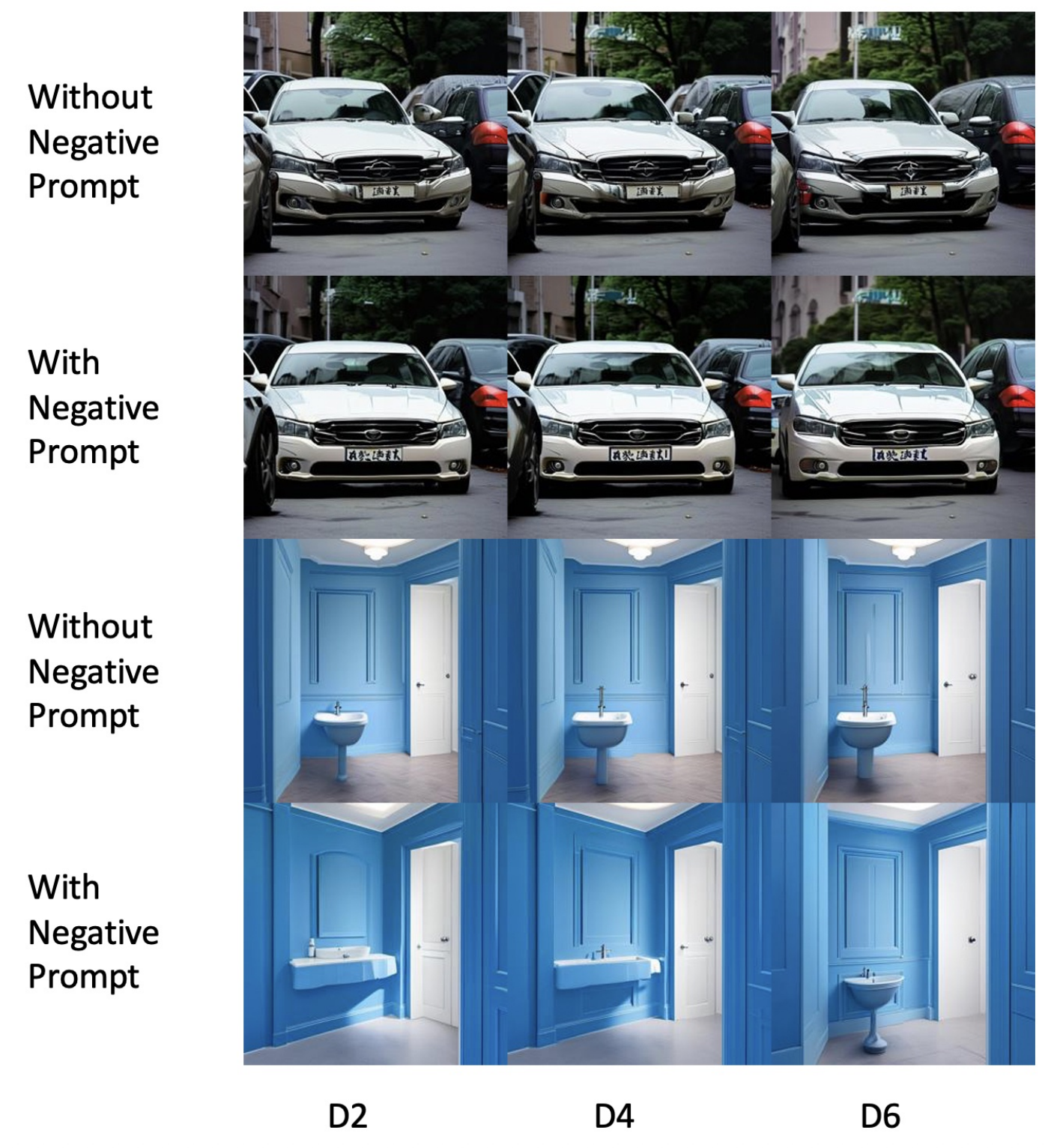}
    \caption{images generated by PixArt-Sigma-XL-1024 at different thresholds with/without negative prompt}
    \label{fig:negative_prompt}
\end{figure}

  Negative conditioning is a technique that widely used to improve generation quality by specifying what to exclude from the generated images. We explored the the impact of negative conditioning on our method using general negative prompt like "Low quality" on PixArt-Sigma-XL. In the case, we found our method can preserve the effect of negative prompt on the generated images as shown in Figure~\ref{fig:negative_prompt}.

% \newpage
% \input{sections/checklist}

\end{document}